%% file: main.tex
\documentclass[11pt]{article}
\usepackage[sc]{mathpazo}
\usepackage{geometry}
\usepackage{color}
\usepackage{array}
\usepackage{bbding}
\usepackage{booktabs}
\usepackage{multirow}
\usepackage{makecell}
\usepackage{multicol}
\usepackage[fleqn]{amsmath}
\usepackage{amssymb,amsthm}
\usepackage{graphicx}
\usepackage{subfig}
\usepackage{natbib}
\usepackage{lscape}
\usepackage{comment}
\usepackage{bm}
\usepackage[title]{appendix}
\usepackage{tabularx}
\usepackage{mathtools}
\usepackage{chngcntr}
\usepackage{authblk}
\usepackage[T1]{fontenc}
\usepackage[utf8]{inputenc}
\usepackage{CJKutf8}  % ✅ 替代 ctex
\usepackage{setspace}
\usepackage{algorithm}
\usepackage{algorithmicx}
\usepackage{algpseudocode}
\usepackage{url}
\usepackage{diagbox}
\usepackage[colorlinks=true, allcolors=blue]{hyperref}
\usepackage{cleveref}
\usepackage{babel}
\usepackage{enumitem}

\usepackage[left, mathlines]{lineno}

\begin{document}\sloppy
	\title{Optimizing Drivers' Discount Order Acceptance Strategies: A Policy-Improved Deep Deterministic Policy Gradient Framework}
	%\title{A Policy-Improved Deep Deterministic Policy Gradient Framework for the Discount Order Acceptance Strategy of Ride-hailing Drivers}
    %\author{Authors’ names blinded for peer review}
	\author{Hanwen Dai\textsuperscript{a}\hspace{2em} Chang Gao\textsuperscript{a}\hspace{2em} Fang He\textsuperscript{a}\footnote{Corresponding author. E-mail address: \textcolor{blue}{fanghe@tsinghua.edu.cn}.}\hspace{2em}Congyuan Ji\textsuperscript{a}\hspace{2em}Yanni Yang\textsuperscript{b}}
	\affil{\small\emph{\textsuperscript{a}Department of Industrial Engineering, Tsinghua University, Beijing 100084, P.R. China}\normalsize}
    \affil{\small\emph{\textsuperscript{b}School of Management and Engineering, Capital University of Economics and Business, Beijing 100084, P.R. China}\normalsize}
	\date{\today}
	\maketitle
%\begin{linenumbers}
	\begin{abstract}
The rapid expansion of platform integration has emerged as an effective solution to mitigate market fragmentation by consolidating multiple ride-hailing platforms into a single application. To address heterogeneous passenger preferences, third-party integrators provide Discount Express service delivered by express drivers at lower trip fares. For the individual platform, encouraging broader participation of drivers in Discount Express services has the potential to expand the accessible demand pool and improve matching efficiency, but often at the cost of reduced profit margins. This study aims to dynamically manage drivers' acceptance of Discount Express from the perspective of an individual platform, incorporating the spatiotemporal demand-supply patterns. The lack of historical data under the new business model necessitates online learning. However, early-stage exploration through trial and error can be costly in practice, highlighting the need for reliable early-stage performance in real-world deployment.
To address these challenges, this study formulates the decision regarding the proportion of drivers accepting discount orders as a continuous control task. In response to the high stochasticity, the opaque matching mechanisms employed by third-party integrator, and the limited availability of historical data, we propose an innovative policy-improved deep deterministic policy gradient (pi-DDPG) framework. The proposed framework incorporates a refiner module to boost policy performance during the early training phase, leverages a convolutional long short-term memory network to effectively capture complex spatiotemporal patterns, and adopts a prioritized experience replay mechanism to enhance learning efficiency. A customized simulator based on a real-world dataset is developed to validate the effectiveness of the proposed pi-DDPG. Numerical experiments demonstrate that pi-DDPG achieves superior learning efficiency and significantly reduces early-stage training losses, enhancing its applicability to practical ride-hailing scenarios. 
	%The increasing adoption of integrated platforms in the ride-hailing industry has introduced new operational complexities, particularly for individual tenants seeking to balance service coverage and profitability under competition. One key decision variable is the proportion of drivers accepting Discount Express orders, which directly influences both match rates and revenue outcomes. This study formulates the drivers’ discount order acceptance setting problem as a continuous control task and proposes a policy-improved Deep Deterministic Policy Gradient (pi-DDPG) framework to optimize tenant-specific control strategies. The proposed framework enhances the conventional DDPG algorithm by introducing a Refiner Module for online action refinement and integrating a Convolutional Long Short-Term Memory (ConvLSTM) network to encode historical spatiotemporal patterns. A custom-built simulator, calibrated with real-world operational data from Beijing, is developed to evaluate algorithm performance. Numerical results show that pi-DDPG achieves superior learning efficiency and higher early-episode rewards compared to baseline DDPG, particularly in sparse-supply and high-variance environments. The proposed approach offers a scalable and adaptive control solution for dynamic decision-making in competitive ride-hailing markets.

		\hfill\break
		
		\noindent\textit{Keywords}: Ride-hailing; Deep Reinforcement Learning; Deep Deterministic Policy Gradient; Online Operations; Online Learning;
		
	\end{abstract}
	\input{data/section1_introduction}
	\input{data/section2_literature}
	\input{data/section3_description}
	\input{data/section4_DDPG}
	\input{data/section5_enhance}
	\input{data/section6_experiment}
	\input{data/section7_conclusion}
%\end{linenumbers}
	\input{data/acknowledgements}\
        \input{data/declaration}
        
	\bibliographystyle{apa}
	\bibliography{ref}
	
	\newpage

\end{document}

%% file: data/section1_introduction.tex
\section{Introduction} \label{sec 1}

The rapid growth of the ride-hailing market has significantly transformed urban transportation, profoundly influencing the daily lives of millions of people worldwide. Driven by the rapid success of ride-hailing services and the relatively low entry barriers, multiple individual platforms now coexist in the local market. As of May 31, 2024, a total of 351 online ride-hailing platforms in China have been granted operating licenses, accompanied by the issuance of approximately 7.033 million driver permits and 2.948 million vehicle operations certificates (\citealp{mot2024}). The intensified platform competition generates two opposing effects. On the one hand, it mitigates the risk of monopolistic dominance, under which one individual platform might maximize profits by charging passengers excessively high fares while simultaneously depressing driver wages below socially efficient levels. On the other hand, excessive competition fragments the market, dispersing passenger demand and driver supply across multiple individual platforms.
Given that ride-hailing markets typically exhibit increasing returns to scale, such fragmentation exacerbates matching frictions, resulting in longer waiting times and greater pickup distances between sparsely distributed drivers and passengers. To alleviate these frictions while preserving the benefits of competition and avoiding inefficient monopolistic dominance, a novel business model -- referred to as \textit{platform integration} -- has emerged, whereby a third-party integrator consolidates multiple individual platforms into a single application (\citealp{zhou2022competition}). For instance, AutoNavi integrates a range of smaller individual platforms, thereby offering passengers a broader set of service options without the need to install multiple platform-specific applications. Through platform integration, passengers requesting rides can be matched with available vehicles from participating individual platforms, while these individual platforms enhance their competitiveness relative to dominant players such as Didi, thereby fostering mutually beneficial outcomes.

To accommodate heterogeneous passenger preferences regarding travel quality and to attract price-sensitive demands, integrators and individual platforms provide a diverse range of service options. Specifically, passengers may choose among Premium services that emphasize comfort and experience, Standard Express services that provide cost-effective transportation, and Discount Express services tailored to price-sensitive users through lower fares. The Discount Express service typically adopts a fixed pricing mechanism, in which fares are predetermined before trip initiation and remain unaffected by real-time traffic conditions or route deviations. Recently, the Discount Express service has attracted substantial demand. For instance, transaction data collected from AutoNavi’s Software-as-a-Service (SaaS) platform indicates that between April 1 and June 30, 2024, 83.72\% of passengers requested the Discount Express service.

\begin{figure}[htbp]
    \centering
    \includegraphics[width=1\linewidth]{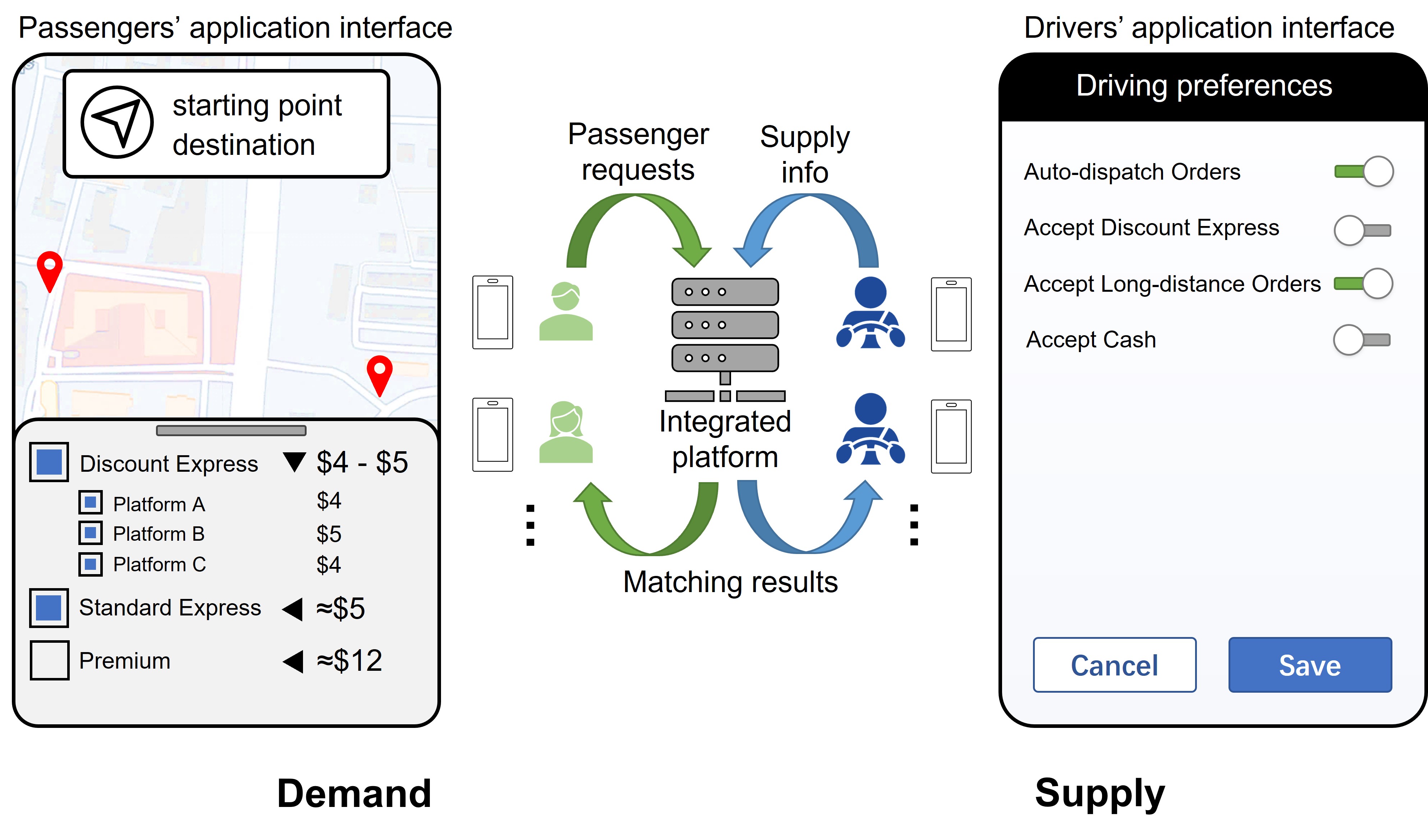}
    \caption{The matching process of passenger demand and driver supply under an integrator}
    \label{fig: ride-hailing service}
\end{figure}

The introduction of multiple service tiers within the third-party integrator substantially complicates the matching process, as illustrated in Figure~\ref{fig: ride-hailing service}. On the demand side, passengers can simultaneously request multiple service types and be matched with drivers affiliated with selected individual platforms, reflecting their preferences regarding price and travel quality. For example, a passenger may concurrently request both Discount Express and Standard Express services from several individual platforms. 
On the supply side, only vehicles meeting higher operational standards are eligible to serve Premium requests. In contrast, all vehicles in the fleet retain the flexibility to decide whether to accept Discount Express requests based on individual preferences and perceived utility. Notably, empirical evidence shows that drivers primarily serving the Premium requests rarely accept Discount Express orders. As a result, these requests are generally fulfilled by drivers who also serve Standard Express requests, ensuring that the lower-priced service maintains a comparable level of quality.
Upon the arrival of new requests, the integrator details the relevant information to the chosen platforms and determines the specific supply-demand assignment by jointly considering drivers' and passengers' preferences. During the assignment process, competition intensifies among individual platforms that are simultaneously chosen by the same passenger. The integrator oversees the overall matching process but withholds its underlying decision logic, leaving the mechanism opaque to the participating individual platforms. Although the integrator coordinates supply–demand matching, it lacks direct authority over driver capacity. In contrast, the participating individual platforms retain operational control by managing drivers’ order acceptance strategies. Accordingly, we designate the individual platforms as the decision-making entities in this study.

\begin{figure}[htbp]
    \centering
    \includegraphics[width=0.8\linewidth]{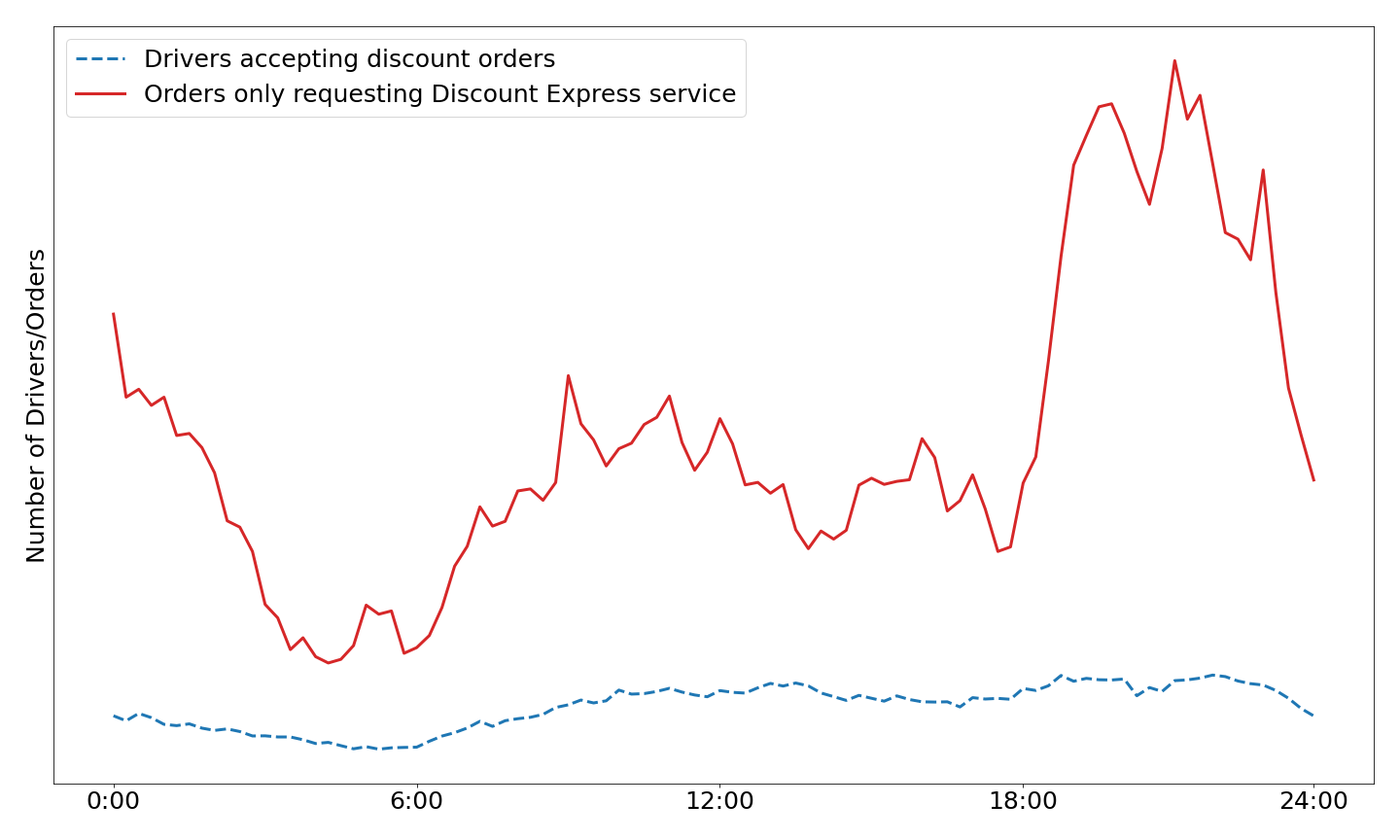}
    \caption{The temporal variation of Discount Express supply-demand in Beijing during a day in October 2023}
    \label{fig: supply and demand in beijing}
\end{figure}

Each service provider faces increasingly complex operational challenges in the aforementioned practical settings. A particularly critical issue lies in efficiently allocating the limited driver supply between Discount Express and Standard Express requests, given that both types of requests are served by the shared supply pool. As illustrated in Figure~\ref{fig: ride-hailing service}, drivers have the flexibility to activate the acceptance of Discount Express in terms of their individual preferences in practice. Upon activating this option, the individual platform simultaneously assigns both Discount Express and Standard Express to drivers, who are then obliged to accept the platform’s matching decisions. However, the lack of complete information may lead drivers to adopt inefficient discount order acceptance strategies, thereby motivating individual platforms to implement centralized control. 
Activating the option to serve Discount Express services represents a double-edged sword for both drivers and individual platforms. For drivers, it expands the potential demand pool and increases matching opportunities at the cost of lower profit margins per trip. For individual platforms, requiring all drivers to accept the Discount Express requests can broaden market reach and strengthen competitive positioning within the integrator.
Nevertheless, this strategy may not be optimal under conditions of supply scarcity. In such scenarios, Standard Express requests are sufficient for drivers, and prioritizing these requests can yield higher earnings for drivers as well as greater profitability for individual platforms. Figure~\ref{fig: supply and demand in beijing} illustrates the variation of demand and supply using operational data in Beijing from a SaaS platform of AutoNavi. Notably, the number of drivers opting into the Discount Express service failed to respond to the surge in demand during the evening peak hours. Thus, it is critical for individual platforms to dynamically determine the optimal number and proportion of drivers to accept discount orders in different supply-demand scenarios. This strategic balance can effectively avoid eroding profit margins while maintaining a competitive edge in the integrator. 

This problem not only necessitates tailored modeling to incorporate supply-demand patterns and inter-platform competition, but also presents substantial technical challenges that must be addressed. The highly dynamism and stochasticity in supply and demand, as well as the black-box matching mechanism in integrator, motivate this study to adopt the deep reinforcement learning framework (DRL). However, the learning efficiency and early-stage exploration cost have become significant barriers for individual platforms to adopt and trust DRL algorithms. 
Platform integration is a newly emerging business model, and individual platforms often lack sufficient historical operational data to support effective pre-training of decision-making policies. Our empirical analysis reveals that more than 90\% of drivers exhibit time-invariant acceptance settings for Discount Express requests. That is, they always either accept or reject Discount Express requests. In consequence, the available data is highly biased and fails to reflect the diversity of potential responses to different control strategies. This data bias further limits the ability of a DRL agent to infer or bootstrap a high-quality policy from offline learning alone, making sample efficiency and safe early-stage online learning critical for real-world adoption. However, in practice, an individual platform often cannot afford the operational losses or service disruptions that may arise during the early learning episodes, when the DRL agent must explore the environment through trial and error and may take suboptimal or even irrational actions. The risk and uncertainty inherent in this exploration phase make direct deployment of online learning strategies particularly costly in real-world settings.

To tackle the aforementioned challenges, this study proposes a customized policy-improved deep deterministic policy gradient (pi-DDPG) framework to develop efficient real-time operational strategies from the individual platforms' perspective, aiming to improve their overall profit and increase drivers' earnings. The proposed strategy enables individual platforms to influence drivers' Discount Express order acceptance behavior by issuing real-time recommendations or, with driver consent, automatically adjusting their acceptance settings. pi-DDPG incorporates a convolutional long short-term memory (ConvLSTM) network to capture the spatiotemporal patterns of supply-demand evolution and further develop adaptive strategies. To improve data efficiency, we adopt a prioritized experience replay (PER) mechanism, which samples historical transitions with higher temporal-difference (TD) errors more frequently, so that the agent can focus on transitions that are expected to yield greater learning progress. Furthermore, we introduce a refiner module that fine-tunes the output actions of the actor network by leveraging critic feedback, thereby improving action quality and convergence stability. Together, these enhancements enable the agent to learn a more accurate and efficient discount order acceptance policy, ultimately supporting individual platforms in achieving a better balance between learning efficiency and exploration cost. The main contributions of this work are summarized as follows.
\begin{itemize}
    \item This study concentrates on an emerging operational problem of coordinating drivers' acceptance of Discount Express requests from an individual platform's perspective under platform integration. Addressing this problem entails managing the trade-off between expanding demand coverage and maintaining profit margins, while also accounting for the dynamic coupling between strategic decision-making and spatio-temporal supply-demand patterns. 
    \item From the technical perspective, we propose a tailored pi-DDPG framework, which extends the conventional deep deterministic policy gradient (DDPG) algorithm by introducing a refiner module into the actor-critic architecture. The refiner adjusts the actor’s output actions through gradient-based feedback from the critic evaluation, enabling online action optimization and improving early-stage performance. To further enhance training efficiency and representation capacity, the framework incorporates the ConvLSTM network to encode the spatio-temporal dynamics of supply-demand evolution and applies PER to improve sampling efficiency. These enhancements collectively improve the learning stability, early-stage performance, and adaptability of the algorithm in complex ride-hailing environments.
    \item We develop a simulator that accurately captures drivers’ online and offline behavior as well as the stochastic nature of order arrivals, and incorporates multi-platform competition mechanism, where the same passenger request may be simultaneously targeted by multiple individual platforms. This enables the evaluation of the proposed pi-DDPG framework under realistic platform-integration settings. Numerical experiments based on a real-world dataset demonstrate that pi-DDPG achieves superior learning efficiency and significantly reduces early-episode training losses, enhancing its applicability to practical ride-hailing scenarios. 
\end{itemize}
%First, we tackle a novel problem scenario by exploring drivers' discount order acceptance strategies under aggregation platforms, an area that has not been extensively studied in the existing literature. Second, we propose a pi-DDPG reinforcement learning framework. Compared to the baseline DDPG algorithm, our enhanced framework demonstrates superior learning efficiency and significantly reduces early-episode training losses, making it more applicable in real-world scenarios. Third, we develop a comprehensive simulator calibrated using real-world operational data from ride-hailing drivers and order data in Beijing. The simulator captures the online-offline behavior patterns of drivers and the stochastic nature of order arrivals. It is specifically designed for demonstration purposes to evaluate the online training performance of the proposed algorithm, providing a controlled yet realistic environment to validate policy effectiveness before real-world deployment.

The remainder of this paper is organized as follows. Section~\ref{sec 2} presents a literature review on ride-hailing platform operations, with a focus on order dispatching under supply-demand imbalances and the reinforcement learning algorithms employed to address these challenges. Additionally, we review different variants of the DDPG algorithm aimed at enhancing its learning efficiency. Section~\ref{sec 3} presents an overview of the problem settings and methodology framework. Section~\ref{sec 4} introduces a baseline DDPG framework for drivers' discount order acceptance control, incorporating spatiotemporal memory and priority experience replay. An extended DDPG framework with a refiner module is then proposed in Section~\ref{sec 5}. Section~\ref{sec 6} outlines the simulator structure and data processing procedure, followed by a presentation of numerical experiment results and a discussion. Finally, Section~\ref{sec 7} offers conclusions and outlines future extensions.

%% file: data/section2_literature.tex
\section{Literature Review} \label{sec 2}

\subsection{Reinforcement learning in assignment strategy of ride-hailing}

Order dispatching and matching represent fundamental operational challenges in ride-hailing services, which demonstrate profound impacts on passenger waiting times, vehicle utilization, and overall platform revenue. A substantial body of literature has recently emerged to address efficient demand-supply assignment, which can be broadly categorized into two methodological frameworks: model-based optimization and reinforcement learning approaches. 
Model-based optimization methods, valued for their interpretability and ease of implementation in real-world scenarios, have attracted considerable attention. 
For instance, \citet{zhang2017taxi} leveraged logistic regression and gradient boosted decision trees to predict driver acceptance behavior, and subsequently applied a tailored hill-climbing algorithm to optimize driver-order assignments, aiming to maximize overall order fulfillment rates. In another line of work, \citet{yang2020optimizing} introduced a spatial probability model to characterize batch matching processes, wherein passenger requests and available drivers are accumulated over short intervals before coordinated allocation. Their theoretical analyses identified optimal matching intervals and radii under various supply-demand scenarios. To address uncertainties inherent in demand emergence, \citet{guo2021robust} employed robust optimization techniques to simultaneously determine vehicle matching and repositioning strategies, aiming to minimize the total vehicle mileage and the number of unsatisfied requests.  Additionally, \citet{gao2024online} presented a two-layer modular modeling framework, where the upper layer strategically manages the spatial transfer of vehicle flows on large timescales to maximize long-term revenues, thereby guiding rapid and real-time vehicle-order matching in the lower layer. \citet{xu2021generalized} proposed a generalized fluid model to effectively capture system dynamics and develop matching strategies, demonstrating its scalability through applications to large-scale instances.

Given the high stochasticity and dynamic nature of ride-hailing systems, reinforcement learning approaches have been increasingly adopted \citep{wang2018deep, tang2019deep, chen2021efficient, sun2022optimizing}. For instance, \citet{chen2021efficient} enhanced Monte-Carlo Tree search for multi-stage dispatching by introducing an effective branch-reduction strategy and a continuous objective function. Comprehensive experiments conducted on both synthetic and real-world city-scale datasets demonstrated substantial improvements in computational efficiency and dispatching performance. \citet{yue2024end} proposed an end-to-end RL framework to optimize the matching process while implicitly integrating driver behavioral predictions. Their approach models the dispatching process as a two-layer Markov decision process (MDP) and employs a novel deep double scalable network to generate effective allocation strategies. \citet{tang2020online} formulated a two-step advisor–student reinforcement learning framework for managing automated electric taxi fleets, integrating policy learning with combinatorial optimization to address charging and dispatching under operational constraints.
To overcome limitations inherent in prior methods — such as short-sighted revenue maximization and neglect of cross-regional driver-flow effects — \citet{yang2024rethinking} proposed a novel goal-reaching collaboration algorithm. This approach features a scoring model that assesses city-wide performance across metrics, including passenger cancellation rates, supply-demand imbalances, and pickup distances, as well as an environmental model predicting future city states. \citet{wang2024reinforcement} employed the discrete choice and Gaussian mixture models to capture passenger and driver heterogeneity, subsequently developing a cooperative DDPG algorithm. This approach integrates individualized neural networks tailored for distinct driver groups with a centralized neural network that coordinates the dispatch objectives across the entire fleet. 

Despite recent advances in improving matching efficiency by incorporating spatiotemporal demand-supply patterns, several critical challenges remain unresolved. First, existing studies primarily focus on strategy development from the perspective of a single platform, overlooking the competitive dynamics among multiple platforms within an integrated platform. Second, the emergence of tiered service offerings -- such as Discount Express and Standard Express -- necessitates more tailored strategies to manage drivers’ acceptance behavior across different order types.

\subsection{DDPG algorithm and its enhancement}

DDPG was first introduced by \citet{lillicrap2015continuous}, which has emerged as a foundational algorithm for continuous control in deep reinforcement learning. To address instability and overestimation during training, two key mechanisms are introduced into the DDPG framework: target networks and experience replay. Target networks help stabilize the learning target by slowly updating separate copies of the actor and critic networks. Experience replay mitigates temporal correlation by storing past transitions in a buffer and sampling them uniformly during training. 

Despite its success, the original DDPG algorithm suffers from issues such as sample inefficiency, instability during training, and sensitivity to hyperparameters. Consequently, a rich body of research has sought to enhance the algorithm along several dimensions. Some research works introduced more efficient sampling methods. \citet{hou2017novel} incorporated the concept of PER into the framework of DDPG learning, which samples transitions based on the magnitude of their temporal-difference error when training the critic network. By emphasizing more informative experiences, PER accelerates convergence and improves sample efficiency. Hindsight experience replay (HER)  allows sample-efficient learning in sparse-reward environments by relabeling stored transitions with alternative goals, such as the achieved goal in the final state of one episode \citep{andrychowicz2017hindsight}.
One prominent line of work involves architectural changes to the DDPG framework. Twin delayed deep deterministic policy gradient (TD3) addresses Q-value overestimation by maintaining two critic networks and using the smaller of the two values for target computation. It also introduces target policy smoothing and delayed policy updates to stabilize training \citep{fujimoto2018addressing}. distributed distributional DDPG (D4PG) integrates distributional value estimation and parallel training to improve efficiency \citep{barth2018distributed}. \citet{lowe2017multi} extended the DDPG algorithm to multi-agent settings by introducing a centralized critic that leverages information about other agents’ policies during training, while preserving decentralized execution. The proposed multi-agent DDPG (MADDPG) significantly improves learning stability and coordination in both cooperative and competitive environments.

Some research explores the potential of incorporating advanced neural network structures to encode more complex system states and enable multi-source data input, thereby improving the performance of the DDPG algorithm. A variety of studies have employed graph neural networks (GNNs) to capture the relational structure of the environment \citep{munikoti2023challenges}. long short-term memory (LSTM) networks have been integrated into the DDPG framework to model temporal dependencies and enhance the agent’s ability to retain and utilize sequential observations \citep{liao2024modelling, li2021lstm, meng2021memory}. More recently, Transformer-based architectures have also been incorporated into multi-agent DDPG frameworks to handle variable-length and multi-source inputs. The attention module at the core of Transformers enables the model to extract important features from complex network state input \citep{guo2024modeling, chen2024transformer}. These neural enhancements significantly improve the perceptual and generalization capabilities of DDPG in complex and dynamic environments.

In light of these research directions, this study proposes a pi-DDPG framework that incorporates two key enhancements. First, a lightweight refiner module is introduced to post-process the actor’s output actions through a local optimization process guided by the critic network. This module refines actions on a per-step basis, effectively bridging the gap between actor approximation and critic value-based evaluation. Second, to enhance the agent's ability to perceive spatiotemporal dynamics, a ConvLSTM-based encoder is embedded into the network architecture. By encoding historical observations over a hexagonal spatial grid structure, the ConvLSTM network captures both temporal dependencies and spatial correlations in the evolving supply–demand environment. These two components jointly improve the learning efficiency and decision accuracy of the DDPG framework in competitive, dynamic ride-hailing scenarios.

%% file: data/section3_description.tex
\section{Problem Settings and Methodology Overview} \label{sec 3}

In this section, we first present the problem settings and assumptions in the dynamic and centralized optimization of the drivers' discount order acceptance problem. We then introduce the overall deep reinforcement learning framework developed to address this problem. 

\subsection{Problem settings and assumptions}

In this paper, we address the problem of managing drivers’ acceptance decisions for Discount Express requests in a spatial network, from the perspective of an individual ride-hailing platform operating under the platform integration.
In the integrated market, a third-party integrator serves as the centralized coordinator that connects passengers and multiple participating individual platforms. Passengers submit their trip requests through the integrator’s mobile application, where they can select their preferred service types (e.g., Standard Express or Discount Express) and choose a set of individual platforms to request from. The integrator collects these real-time passenger demands together with the available driver information from multiple participating individual platforms and performs the order–driver matching. Each individual platform, in turn, receives the demand information of passengers who have selected it, and retains control over its own fleet management and drivers’ Discount Express order acceptance settings. The integrator’s matching mechanism is treated as a black box that is opaque to the individual platforms. Meanwhile, passengers and drivers constitute the external environment of the system -- their arrivals, cancellations, and online/offline behaviors evolve stochastically and provide the dynamic context within which the individual platforms and the integrator operate.

\begin{figure}[htbp]
	\begin{minipage}{0.48\textwidth}
		\centering
		\includegraphics[width=\linewidth]{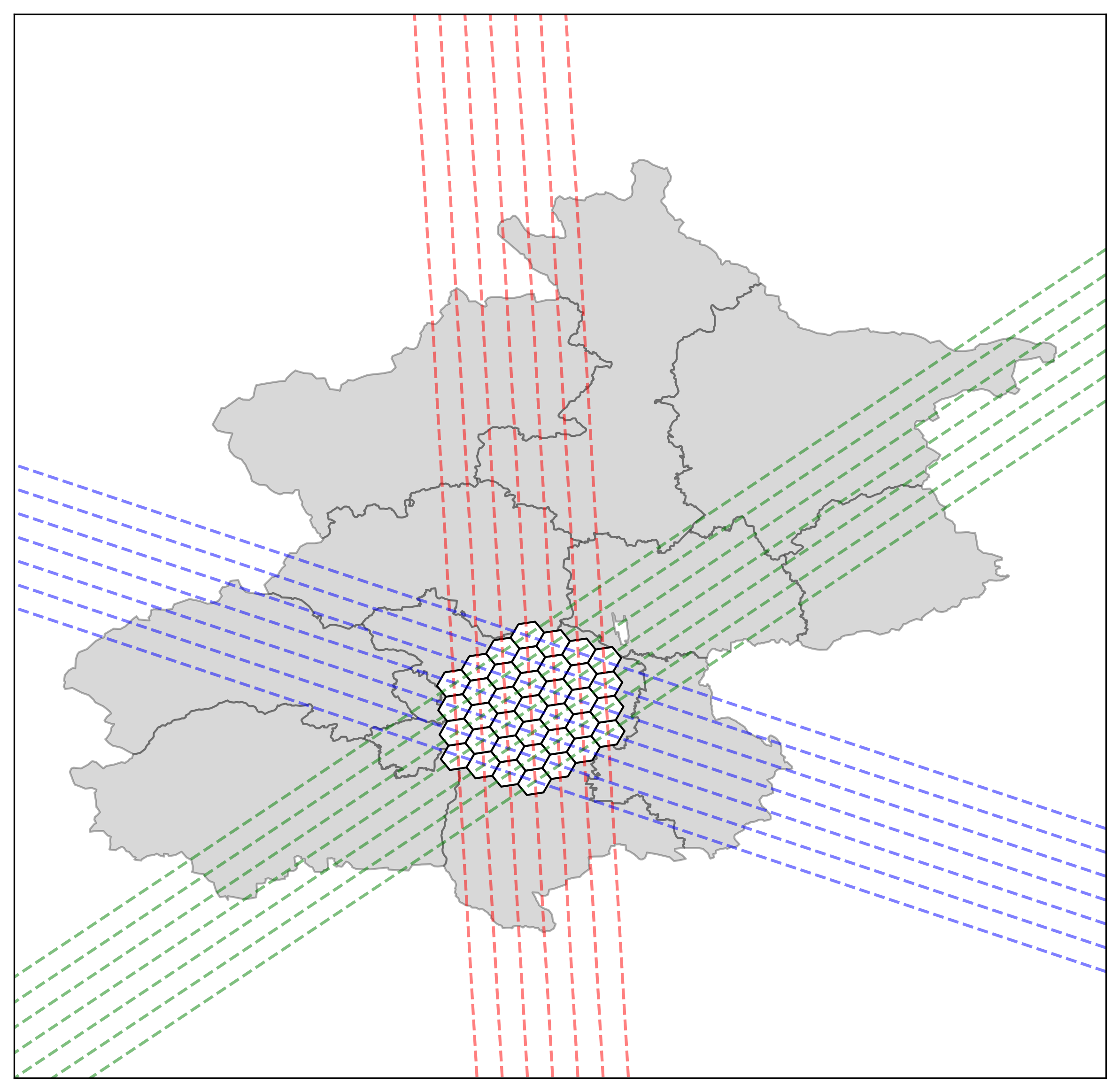}
	\end{minipage}
	\begin{minipage}{0.48\textwidth}
		\centering
		\includegraphics[width=\linewidth]{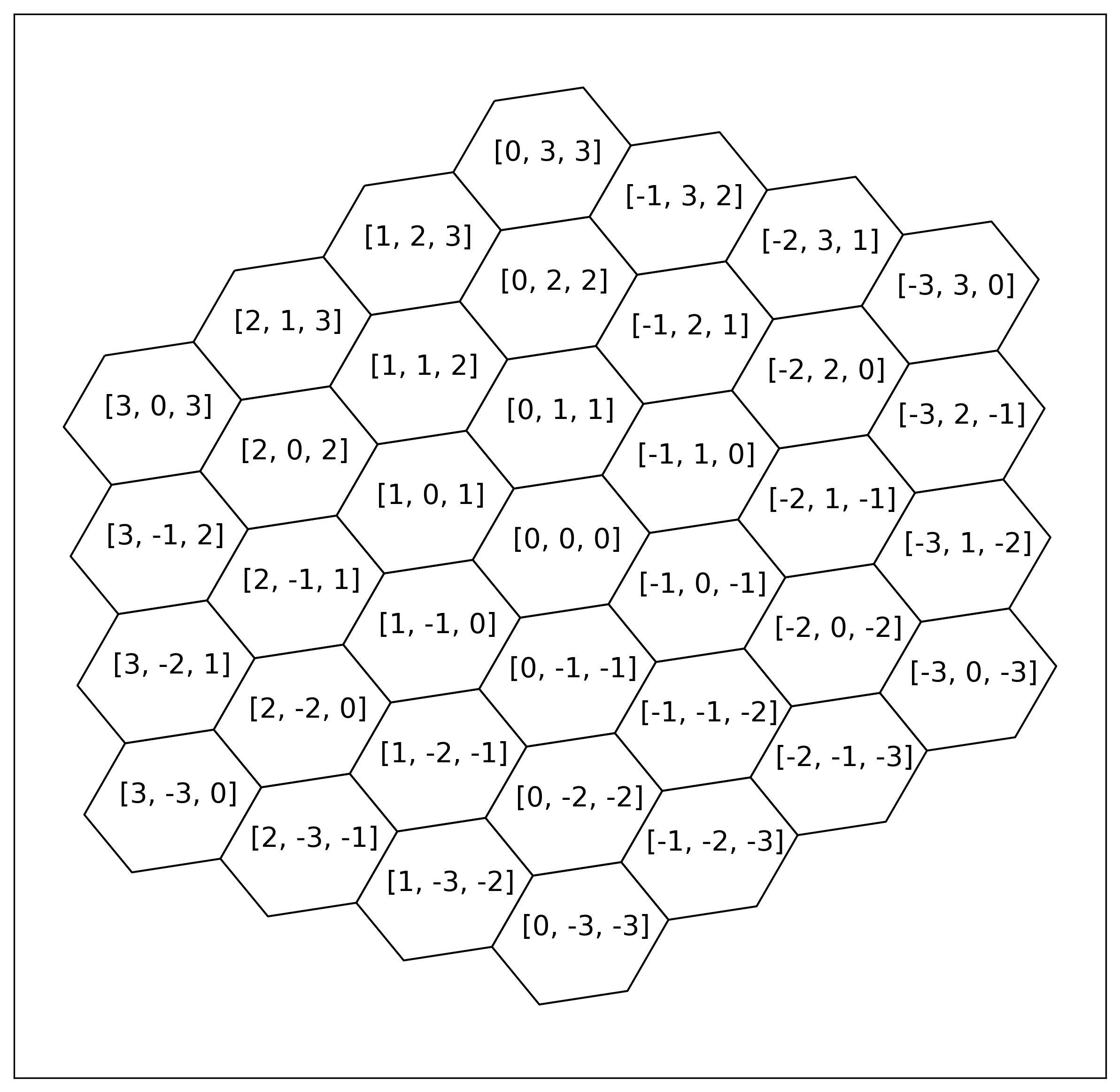}
	\end{minipage}
    \caption{Hexagonal grid network with 3D matrix encoding} 
	\label{fig: hexagonal grid encoding}
\end{figure}

\begin{figure}[htbp]
    \centering
    \includegraphics[width=1\linewidth]{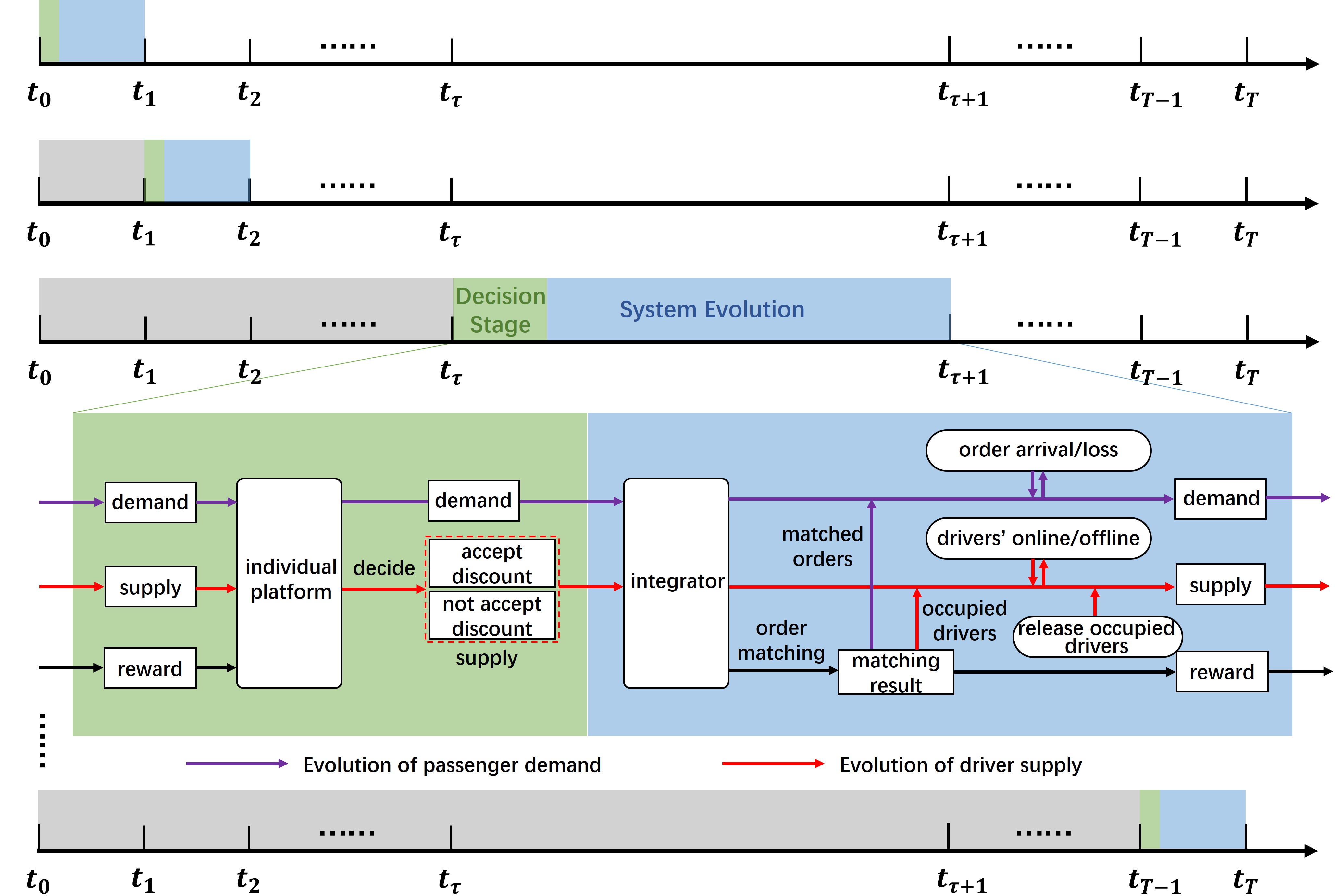}
    \caption{Operational framework of drivers' discount order acceptance setting control}
    \label{fig: operational framework}
\end{figure}

The time horizon is discretized into $T$ time intervals of equal length, and the target area is partitioned into a hexagonal grid network, as illustrated in Figure~\ref{fig: hexagonal grid encoding}, where each grid is encoded using three-dimensional coordinates. Figure~\ref{fig: operational framework} illustrates the operational framework of the individual platform. At the beginning of each interval, the individual platform determines the proportion of drivers in each grid that are capable of serving Discount Express requests, aiming to enhance profitability and maintain competitiveness against other individual platforms. We assume that drivers are fully compliant with the individual platform’s guidance, motivated by incentive mechanisms offered by the individual platform.
Based on this proportion, the individual platform activates the option for a subset of drivers with the strongest revealed preference for Discount Express requests, while keeping it disabled for the remainder. This rule balances system-wide efficiency with the preservation of driver experience. Drivers’ preferences are inferred from their historical acceptance frequency; however, the precise estimation of such preferences lies beyond the scope of this study and is left for future research. Importantly, the individual platform operates with incomplete information: they do not observe the real-time driver distribution or acceptance strategies of competitors. 

During each interval, the operational process can be divided into two interconnected stages, as illustrated in Figure \ref{fig: operational framework}.
In the Decision Stage (green area), the individual platform determines the proportion of drivers in each grid that enable the Discount Express option. This decision is made based on the current demand–supply information and past matching outcomes. Once the decision is implemented, the resulting configuration of drivers -- some accepting and others rejecting discount orders -- forms the input of driver supply for the following stage.
Subsequently, the System Evolution Stage (blue area) begins. During this period, the integrator conducts order–driver matching by combining the requests submitted by passengers with the available driver pool across different individual platforms. Although the integrator’s matching logic remains opaque to individual platforms to preserve fairness, it may, for example, assign each request to the nearest available driver or to the driver with the highest credit score within a certain radius. After matching, drivers fulfill their assigned trips, while the system continues to evolve stochastically through: (1) stochastic order arrivals, (2) expiration of unmatched orders, (3) driver online/offline behavior, and (4) trip completions with occupied drivers released. At the end of the interval, the realized matching outcomes and reward signals -- reflecting platform performance, demand satisfaction, and driver utilization -- are aggregated as the environment feedback. These outcomes define the new state that will serve as the input for the next Decision Stage at time $t_{\tau + 1}$, thereby completing one full decision–evolution cycle. This iterative process continues throughout the entire time horizon, allowing individual platforms to adapt their drivers' discount order acceptance strategies in response to continuously changing demand and supply dynamics.

%During each interval, newly-emerging passengers submit their requests to the integrator, specifying their preferred service types and selected platforms. The integrator then forwards the relevant information to the chosen platforms and performs the matching process using the collective driver pool. The exact matching logic remains a black-box to individual platforms to ensure fairness. For instance, the integrator can assign the request to the nearest available driver, or to the driver with the highest credit within a predetermined radius. Platforms subsequently receive the matching outcomes.
%The system continues to evolve within the interval through the drivers' fulfillment of assigned passenger orders, drivers' stochastic online and offline transitions, and stochastic order arrivals in each grid. 
%As the time horizon progresses, the decision-making process as depicted above will be repeated, with platforms continuously updating their decisions on drivers' acceptance towards Discount Express based on historical matching outcomes and demand-supply information.  

\subsection{Methodology overview}

\begin{figure}[htbp]
    \centering
    \includegraphics[width=1\linewidth]{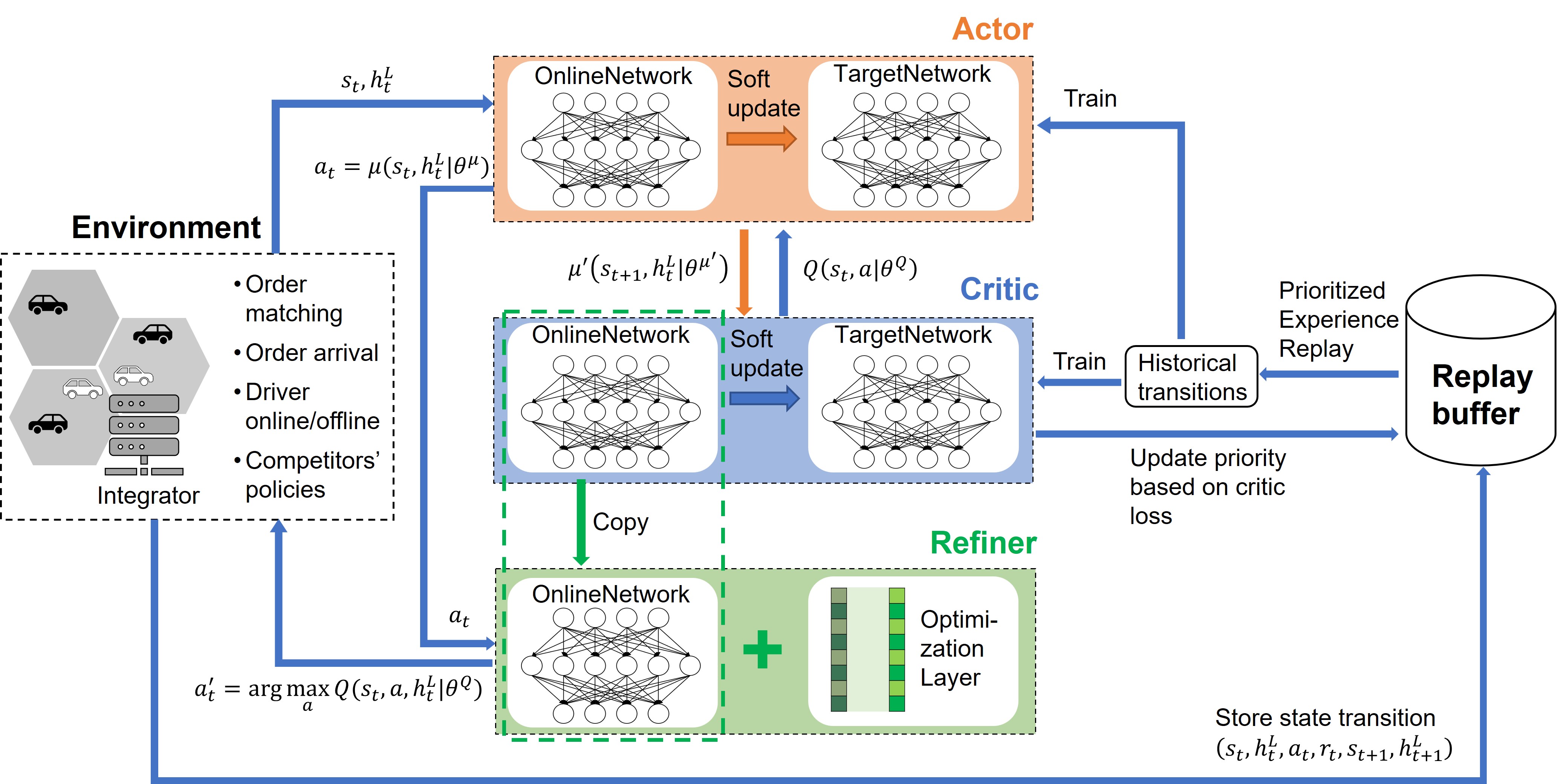}
    \caption{The overall framework of pi-DDPG}
    \label{fig: DDPG framework}
\end{figure}

In our framework, we leverage the DDPG algorithm to enable intelligent control over drivers’ discount order acceptance. As illustrated in Figure~\ref{fig: DDPG framework}, the methodology involves the interaction of five key components: the environment, the actor, the critic, the optimizer, and the replay buffer.

The environment encompasses the integrator, multiple competing individual platforms, and stochastic passenger demand and driver supply. The integrator serves as the centralized system that collects and integrates real-time supply and demand information, performs order matching, and mediates interactions among drivers, passengers, and individual platforms. Passenger requests arrive randomly over time and space, introducing uncertainty into the decision-making process. Meanwhile, multiple individual platforms compete for a shared pool of orders, each adjusting its control policies to improve order fulfillment. This complex and dynamic environment forms the basis on which the learning agent, a single individual platform, must adapt its strategy in response to real-time supply-demand fluctuations and competitive pressure from other platforms.

The learning framework is designed from the perspective of an individual platform, which independently maintains its own actor, critic, refiner, and replay buffer modules. The \textbf{critic} network estimates the action-value function and provides gradient-based feedback to guide the training of the actor. Conditioned on the current system state and memory information, the \textbf{actor} network generates a deterministic action that specifies the proportion of drivers accepting Discount Express orders in each spatial grid. Given the initial action proposed by the actor, the \textbf{refiner} module performs online action optimization by leveraging gradient feedback from the critic, thereby enhancing action quality before execution. Both the actor and critic are equipped with target networks that are softly updated to improve learning stability. To further enhance sample efficiency, a prioritized experience \textbf{replay buffer} is used to store past transitions, where samples with higher critic losses from previous learning epochs are assigned greater sampling priorities during training. The detailed architecture and training procedure of the baseline DDPG and the enhanced pi-DDPG framework will be discussed in Section~\ref{sec 4} and Section~\ref{sec 5}.

%% file: data/section4_DDPG.tex
\section{Fundamentals of DDPG Algorithm} \label{sec 4}

\subsection{Problem statement}
We formulate the multi-period driver order-acceptance process for each individual platform as a MDP, defined by a tuple $(\mathcal S, \mathcal A, \mathcal P, \mathcal R, \gamma)$, where $\mathcal S$ is the state space, $\mathcal A$ is the action space, $\mathcal P$ denotes the transition dynamics, $\mathcal R$ is the reward function, and $\gamma$ is the discount factor for future rewards.

\textbf{Agent.} Each individual platform is modeled as an independent agent. An individual platform makes decisions solely based on its own operational status, without access to the information of other individual platforms, including their driver availability or order fulfillment results. And individual platforms cannot interfere with others’ discount order acceptance strategies.

\textbf{State}. At decision epoch $t$, each individual platform observes only its own operational status, including the spatial distribution of its available drivers and the passenger orders specifically requesting its service. Let $\mathcal I$ denote the set of hexagonal grids, and $\mathcal J$ denote the set of order types. Formally, the state observed by an individual platform is represented as follows:
\begin{itemize}
    \item The \textbf{platform-specific order pool} is denoted as $\bm{d}_{t} = \{d_{i,j,t}\}_{i \in \mathcal{I}, j \in \mathcal{J}}$, where $d_{i,j,t}\in\mathbb N$ indicates the number of Type-$j$ orders located in grid $i$ at time $t$ that request service from that platform.
    \item The \textbf{platform-specific driver pool} is denoted as $\bm{n}_{t} = \{n_{i,t}^+, n_{i,t}^-\}_{i \in \mathcal{I}}$, where $n_{i,t}^+, n_{i,t}^-\in \mathbb N$ represent the numbers of drivers in grid $i$ who accept and do not accept discount orders, respectively, and are affiliated with the platform.
\end{itemize}
The complete observable state for the individual platform is then constructed by concatenating the two components:
\begin{linenomath}
\begin{equation}
    \bm s_t = [\bm{n}_{t}, \bm{d}_{t}] \in \mathcal{S}
\end{equation}
\end{linenomath}
This limited observability reflects the decentralized nature of platform operations on the integrated platform, where each platform acts independently and has no access to global supply-demand information.

\textbf{Action}. The action $\bm{a}_t = \{a_{i,t}\}_{i\in \mathcal{I}} \in \mathcal A$ represents the proportion of drivers in each grid $i$ that are instructed to accept Discount Express orders. Each element $a_{i,t} \in [0,1]$ controls the discount-setting intensity spatially.

\textbf{Reward}. 
At the end of each time step $t$, the individual platform receives detailed matching results from the integrated platform, denoted as $\bm{b}_t = \{\bm{b}_{i,t}\}_{i \in \mathcal{I}}$, where each $\bm{b}_{i,t}$ represents the set of matched driver–order pairs in grid $i$. Based on this information, the individual platform can directly compute its immediate reward $r_t$ as the total service fees collected from orders fulfilled by its drivers. Specifically, Discount Express orders yield lower revenue compared to Standard Express orders.

\textbf{State transition.}
The state of each individual platform evolves according to its current state $\bm s_t$, action $\bm a_t$, and a set of exogenous processes. After the individual platform selects $\bm a_t$, these processes—including integrator’s order-matching rule, stochastic order arrivals, driver online/offline dynamics, and competitors’ policies—jointly determine the matching outcomes and the next state. This evolution is represented by the transition kernel $P(\bm s_{t+1}\mid \bm s_t,\bm a_t;\Theta)$, where $\Theta$ collects all exogenous components. The individual platform’s RL agent does not directly observe or control these exogenous processes but adapts its policy $\pi(\bm a|\bm s)$ through repeated interaction with the environment.

\textbf{Objective.} The agent's learning goal is to derive a policy $\pi: \mathcal S \rightarrow \mathcal A$ that maximizes the expected cumulative discounted reward $\mathbb E\Big[\sum_{t=0}^T\gamma^tr_t|\pi\Big]$ over time.

\subsection{DDPG algorithm}

The DDPG algorithm is an actor–critic, model-free, off-policy reinforcement learning method designed for environments with continuous action spaces. It combines the strengths of deterministic policy gradient and Q-learning, enabling scalable and stable learning in high-dimensional control tasks. The fundamental components and concepts of the DDPG algorithm are described as follows:

\textbf{Actor.} The actor network deterministically maps each observed state to an action, representing the current policy $\pi$ of the agent. It is defined as a function $\mu(\bm s|\bm \theta^\mu): \mathcal{S} \rightarrow \mathcal{A}$, where $\bm \theta^\mu$ are the trainable parameters of the actor network.

\textbf{Critic.} The critic network estimates the Q-value function $Q^\pi(\bm s, \bm a|\bm \theta^Q): \mathcal{S} \times \mathcal{A} \rightarrow \mathbb{R}$, which evaluates the expected cumulative discounted reward starting from state $\bm s_t$ and action $\bm a_t$, following policy $\pi$. Here, $\bm \theta^Q$ are the trainable parameters of the critic network. Ideally, with sufficient training, the critic converges to:
\begin{linenomath}
\begin{equation}
    Q^\pi(\bm s_t,\bm a_t|\bm \theta^Q) = \mathbb E\Big[\sum_{i=t}^{T}\gamma^{i-t}r_{i}|\pi\Big]
\end{equation}
\end{linenomath}
For notational simplicity, we omit the superscript $\pi$ in the remainder of the paper and use $Q(\bm s_t,\bm a_t|\bm \theta^Q)$ to denote the action-value function under the current actor's policy.

\textbf{Target networks.} To improve training stability, DDPG maintains slowly-updated copies of both the actor and critic networks, denoted as $\mu'(\bm s|\bm \theta^{\mu'})$ and $Q'(\bm s, \bm a|\bm \theta^{Q'})$. These target networks are updated using a soft update rule:
\begin{linenomath}
\begin{equation}
\bm \theta^{Q'} \leftarrow \tau\bm \theta^Q + (1-\tau)\bm \theta^{Q'}, \quad \bm \theta^{\mu'} \leftarrow \tau\bm \theta^\mu + (1-\tau)\bm \theta^{\mu'} \label{equ: soft update}
\end{equation}
\end{linenomath}
where $\tau \in (0, 1)$ controls the update rate. Target networks prevent the critic from making large, destabilizing updates in early stages of training.

\textbf{Replay buffer.} DDPG employs an experience replay buffer $\mathcal{R}$ to store transition tuples $(\bm s_i, \bm a_i, r_i, \bm s_{i+1})$. During training, a mini-batch of size $N$ is randomly sampled from $\mathcal{R}$ to break the temporal correlation between samples and improve sample efficiency. These sampled transitions form the basis for updating both the critic and actor networks.

\textbf{Actor update.} The actor network is updated using the deterministic policy gradient:
\begin{linenomath}
\begin{equation}
\nabla_{\bm \theta^\mu} J \approx \frac{1}{N} \sum_{i=1}^{N} \nabla_a Q(\bm s,\bm a|\bm \theta^Q)\big|_{\bm s=\bm s_i,\bm a=\mu(s_i)} \cdot \nabla_{\bm \theta^\mu} \mu(\bm s|\bm \theta^\mu)\big|_{\bm s=\bm s_i} \label{equ: policy gradient}
\end{equation}
\end{linenomath}
This gradient update encourages the actor to choose actions that yield higher Q-values as estimated by the critic.

\textbf{Critic update.} Given the sampled transition $(\bm s_i, \bm a_i, r_i, \bm s_{i+1})$, the critic is trained by minimizing the mean squared temporal-difference (TD) error $\delta_i$ between its Q-value estimate and a target Q-value computed using the target networks:
\begin{linenomath}
\begin{equation}
L(\bm \theta^Q) = \frac{1}{N} \sum_{i=1}^{N} \delta_i^2 = \frac{1}{N} \sum_{i=1}^{N}\left(y_i - Q(\bm s_i, \bm a_i|\bm \theta^Q)\right)^2 \label{equ: critic loss}
\end{equation}
\end{linenomath}
with target Q-value defined as:
\begin{linenomath}
\begin{equation}
y_i = \begin{cases}
    r_i + \gamma Q'(\bm s_{i+1}, \mu'(\bm s_{i+1}|\bm \theta^{\mu'})|\bm \theta^{Q'}), & \text{if } i<T\\
    r_i & \text{otherwise}
\end{cases} \label{equ: target Q value}
\end{equation}
\end{linenomath}

\textbf{Episode reward.} An \textbf{episode} is referred as a complete sequence of interactions between the learning agent and the environment from an initial state to an end state, $\{(\bm s_t, \bm a_t, r_t, \bm s_{t+1})\}_{t\in\{0,\cdots,T\}}$.
The episode reward is defined as the cumulative reward collected over an entire episode, $\sum_{t\in\{0,\cdots,T\}}r_t$. It serves as a \textbf{key performance metric} to evaluate the effectiveness of the agent’s policy during training.

\subsection{Leveraging Memory of Spatiotemporal Information}

To enhance the decision-making capability of the DDPG framework in a dynamic and spatially complex environment, we extend the state representation by incorporating historical spatiotemporal memory. Specifically, in addition to the current state $s_t$, we define a memory input $\bm h_t^L$ that encodes information from the preceding $L$ steps within the same episode. This memory includes each time step’s driver distribution, passenger demand, platform's action, and matching results:
\begin{linenomath}
\begin{equation}
    \bm h_t^L = [\bm n_{t-L}, \bm d_{t-L}, \bm a_{t-L}, \bm b_{t-L}, \cdots, \bm n_{t-1}, \bm d_{t-1}, \bm a_{t-1}, \bm b_{t-1}]
\end{equation}
\end{linenomath}
If the current step $t < L$, the remaining entries in $\bm h_t^L$ are zero-padded. The input of historical memory allows the policy network to access short-term spatiotemporal patterns and enhances its responsiveness to evolving supply-demand dynamics.

To encode spatial information more effectively, the target area is partitioned into hexagonal grids. Hexagons provide a symmetric and uniform definition of neighborhood connectivity, with consistent distance properties and lower edge-to-area ratios compared to square grids. These advantages help mitigate edge-related biases and better preserve inflow and outflow patterns across locations. Due to these advantages, hexagons are adopted as calculation units by many organizations, such as H3 grid indexing system developed by Uber (\citealp{uberh3}).
Each historical snapshot is organized as a three-dimensional matrix $\mathcal{X}_{t'}$ for each $t' \in \{t-L, \cdots, t-1\}$, where each matrix element corresponds to a hexagonal grid. Suppose grid $i$ has spatial coordinates $(x_i, y_i, z_i)$ under hexagonal indexing, then the corresponding element in $\mathcal{X}_{t'}$ stores the historical driver state, order state, executed action, and matching result in grid $i$ at time $t'$. The complete memory input $\bm h_t^L$ is thus transposed into a sequence of 3D tensors $\{\mathcal{X}_{t-L}, \cdots, \mathcal{X}_{t-1}\}$.

To extract meaningful spatiotemporal features from this structured input, we employ a ConvLSTM network (\citealp{shi2015convolutional}). ConvLSTM is an extension of traditional LSTM where fully connected transformations are replaced by convolutional operations. This architectural change enables ConvLSTM to simultaneously capture temporal dependencies and spatial correlations in grid-based input. Its recurrent structure allows temporal memory propagation, while the convolutional kernel ensures local spatial interactions are retained at each time step. As a result, ConvLSTM is well-suited for modeling supply-demand dynamics that evolve over both time and space. Through the integration of hexagonal spatial encoding and ConvLSTM-based temporal embedding, our framework captures both the geographic structure and dynamic evolution of the system state. This enriched state representation serves as the input to both actor and critic networks in the DDPG framework, enhancing the model’s ability to generate high-quality, context-aware decisions.
 
\begin{figure}[htbp]
    \centering
    \includegraphics[width=0.8\linewidth]{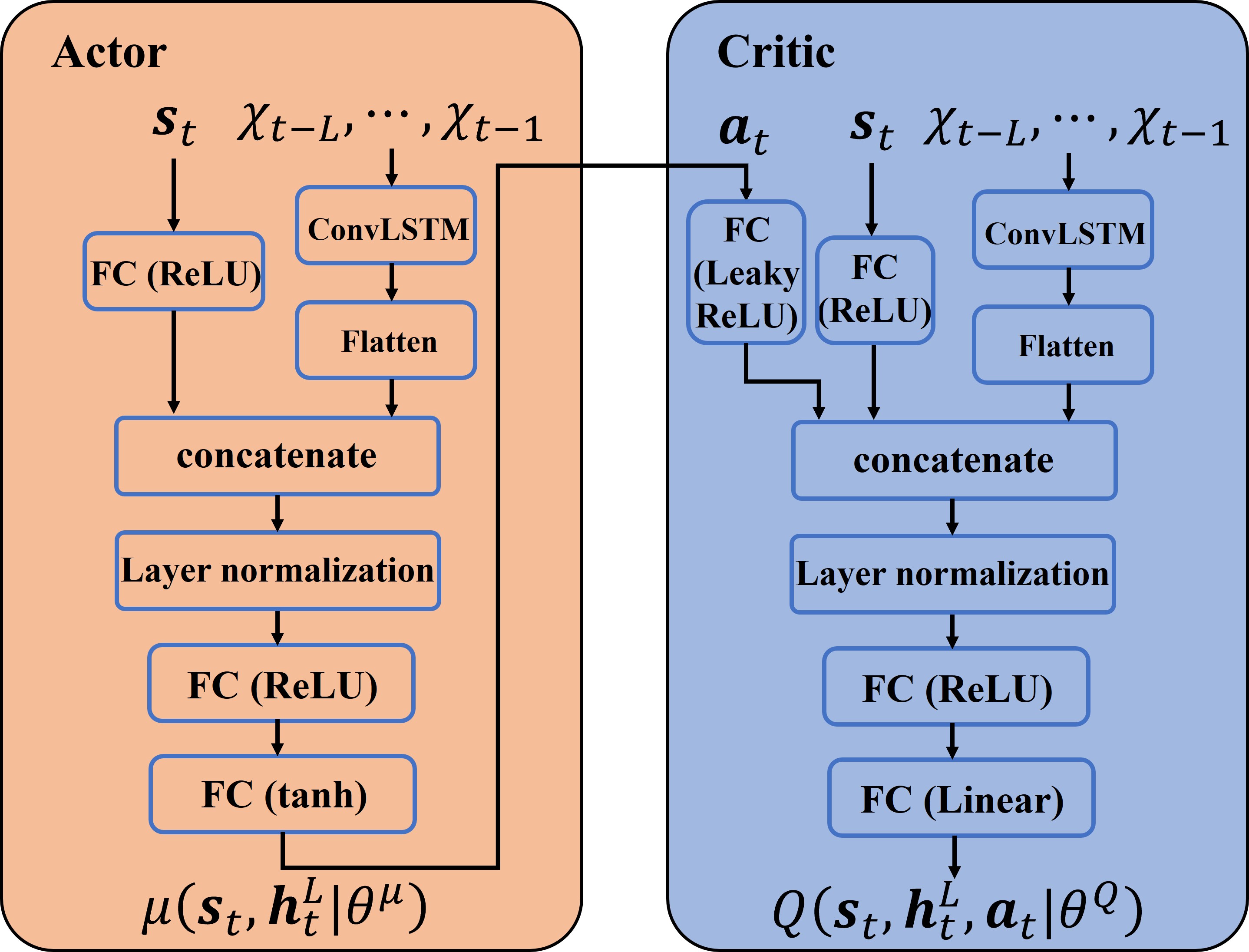}
    \caption{The network structure with the spatiotemporal memory input}
    \label{fig: network structure}
\end{figure}

The introduction of the spatiotemporal memory $\bm h_t^L$ and embedding of the ConvLSTM lead to structural modifications of the original DDPG framework, as shown in Figure~\ref{fig: network structure}. Specifically, the input to both actor and critic networks is augmented from the current state $\bm s_t$ to the enriched state-memory pair $(\bm s_t, \bm h_t^L)$. As a result, the training samples stored in the replay buffer are extended to include memories, i.e., each historical transition tuple becomes $(\bm s_t, \bm h_t^L, \bm a_t, r_t, \bm s_{t+1}, \bm h_{t+1}^L)$. Furthermore, with a sampled batch of transitions, the optimization process of both actor network $\mu(s, \bm h^L|\bm \theta^\mu)$ and critic network $Q(\bm s, \bm h^L,\bm a|\bm \theta^Q)$ is adapted as:
\begin{linenomath}
\postdisplaypenalty=0
\begin{align}
& \min_{\bm \theta^Q}L(\bm \theta^Q) = \frac{1}{N} \sum_{i=1}^{N} \left(y_i - Q(\bm s_i, \bm h_i^L, \bm a_i|\bm \theta^Q)\right)^2 \label{equ: critic loss with memory}\\
& y_i = \begin{cases}
    r_i + \gamma Q'(\bm s_{i+1}, \bm h_{i+1}^L, \mu'(\bm s_{i+1}, \bm h_{i+1}^L|\bm \theta^{\mu'})|\bm \theta^{Q'}), & \text{if } i<T\\
    r_i & \text{otherwise}
\end{cases} \label{equ: target Q value with memory}\\
& \nabla_{\bm \theta^\mu} J \approx \frac{1}{N} \sum_{i=1}^{N} \nabla_a Q(\bm s,\bm h^L, \bm a|\bm \theta^Q)\big|_{\bm s=\bm s_i, \bm h^L = \bm h_i^L, \bm a=\mu(\bm s_i, \bm h_i^L)} \cdot \nabla_{\bm \theta^\mu} \mu(\bm s,\bm h^L|\bm \theta^\mu)\big|_{\bm s=\bm s_i, \bm h^L = \bm h_i^L} \label{equ: policy gradient with memory}%
\end{align}
\end{linenomath}

\subsection{Priority Experience Replay}

To enhance the efficiency of training batch sampling from historical transitions, we adopt the Priority Experience Replay (PER) mechanism (\citealp{hou2017novel}). The historical transitions stored in the replay buffer are ranked based on their TD errors, where a higher rank, $rank(j)$, for transition $j$ indicates a greater absolute TD error, $|\delta_j|$. The probability of selecting a historical transition $j$ is given by
\begin{linenomath}
\begin{equation}
    P(j) = \frac{D_j^\alpha}{\sum_k D_k^\alpha}
\end{equation}
\end{linenomath}
where $D_j = \frac{1}{rank(j)} > 0$, and parameter $\alpha$ controls the degree of prioritization. To implement this PER framework, we maintain a table that records the rank of each transition based on its TD-error from previous training epochs. The PER mechanism prioritizes transitions with higher TD-errors, as they offer greater learning potential, contributing to the reduction of training loss and the improvement of the critic network's evaluation accuracy. However, transitions with lower TD-errors still have a chance of being replayed, ensuring diversity in the sampled transitions.

To stabilize the training process, importance-sampling weights are introduced. The training weight of a historical transition $j$ is defined as
\begin{linenomath}
\begin{equation}
    w_j = \frac{1}{S^\beta \cdot P(j)^\beta} \label{equ: importance-sampling weights}
\end{equation}
\end{linenomath}
where $S$ is the size of the replay buffer, and the parameter $\beta$ controls the extent of the correction applied. These weights mitigate the bias caused by prioritized sampling, ensuring that the training process remains stable and unbiased while still benefiting from the PER mechanism. Consequently, a refined training loss function for the critic network is formulated as
\begin{linenomath}
\begin{equation}
    L(\bm \theta^Q) = \frac{1}{N}\sum_j w_j\delta_j^2. \label{equ: weighted loss}
\end{equation}
\end{linenomath}

The complete DDPG algorithm for dynamic control of drivers' discount order acceptance setting is presented in Algorithm~\ref{algo: DDPG}.

\begin{algorithm}[ht!]
	\caption{DDPG for Drivers' Discount Order Acceptance Setting Control}
	\label{algo: DDPG}
	\begin{algorithmic}[1]
	\State Initialize critic $Q(\bm{s}, \bm{h}^L, \bm{a}|\bm \theta^Q)$, actor $\mu(\bm{s}, \bm{h}^L|\bm \theta^\mu)$, and their target networks $Q'$, $\mu'$
	\State Initialize prioritized replay buffer $\bm{R}$ and Gaussian noise $\mathcal N(0,\sigma^2)$
	\For{each episode $i \in \{1,\cdots,K\}$}
	    \State Observe initial state $\bm{s}_0$ and historical latent state $\bm{h}_0^L$
	    \For{each time step $t \in \{0, \cdots, T\}$}
	        \State Obtain action $\bm{a}_t = \mu(\bm{s}_t, \bm{h}_t^L|\bm \theta^\mu)$
	        \State Execute the action with noise for exploration $\bm{a}_t \leftarrow \bm{a}_t + \mathcal N(0, \sigma^2)$
            \State Observe reward $r_t$, next state $\bm{s}_{t+1}$ and $\bm{h}_{t+1}^L$
	        \State Store transition $(\bm{s}_t, \bm{h}_t^L, \bm{a}_t, r_t, \bm{s}_{t+1}, \bm{h}_{t+1}^L)$ in prioritized buffer $\bm{R}$, ranked by $|\delta_t|$
	        \State Sample a minibatch of $N$ transitions $\{(\bm{s}_j, \bm{h}_j^L, \bm{a}_j, r_j, \bm{s}_{j+1}, \bm{h}_{j+1}^L)\}_{j=1}^N$ from $\bm{R}$ with PER
	        \State Compute TD error $\delta_j = y_j - Q(\bm{s}_j, \bm{h}_{j}^L, \bm{a}_j|\bm \theta^Q)$
	        \State Compute importance-sampling weights for the sampled mini-batch by Equation~\ref{equ: importance-sampling weights}
            \State Update sampled transition $j$'s rank according to its absolute TD-error $|\delta_j|$
	        \State Update critic $Q(\bm{s}, \bm{h}^L, \bm{a}|\bm \theta^Q)$ by minimizing the weighted loss by Equation~\ref{equ: weighted loss}
	        \State Update actor $\mu(\bm{s}, \bm{h}^L|\bm \theta^\mu)$ using the policy gradient by Equation~\ref{equ: policy gradient}
	        \State Soft update target networks' weights $\bm \theta^{Q'}, \bm \theta^{\mu'}$ by Equation~\ref{equ: soft update}
	    \EndFor
        \State Decline $\sigma$ of the Gaussian noise
	\EndFor
	\end{algorithmic}
\end{algorithm}

%% file: data/section5_enhance.tex
\section{pi-DDPG Algorithm} \label{sec 5}

To further enhance the performance of our DDPG-based drivers' discount order acceptance setting control framework, we extend the standard DDPG architecture with a refiner module designed for online action refinement. 
The following subsections detail the module structure, online refinement algorithm, and the complete pi-DDPG algorithm for the control problem.

\subsection{Refiner module structure}

Although the actor network is trained to approximate the optimal policy, its outputs can be suboptimal due to function approximation limitations, unstable gradient feedback, insufficient exploration, or bad sampling. These issues are especially pronounced in dynamic, competitive environments like a multi-platform ride-hailing market, leading to poor performance of DDPG algorithm during the early episodes of learning.

To address this, our improved framework incorporates a lightweight post-processing step that refines the actor's output by consulting the critic's value estimates. During the online operations, rather than relying solely on the actor’s direct decision, the final action is selected by exploring alternatives in its local neighborhood and identifying one with a higher estimated Q value. This refinement serves as a targeted correction that enhances policy quality without incurring significant computational overhead.

\begin{figure}[htbp]
    \centering
    \includegraphics[width=0.75\linewidth]{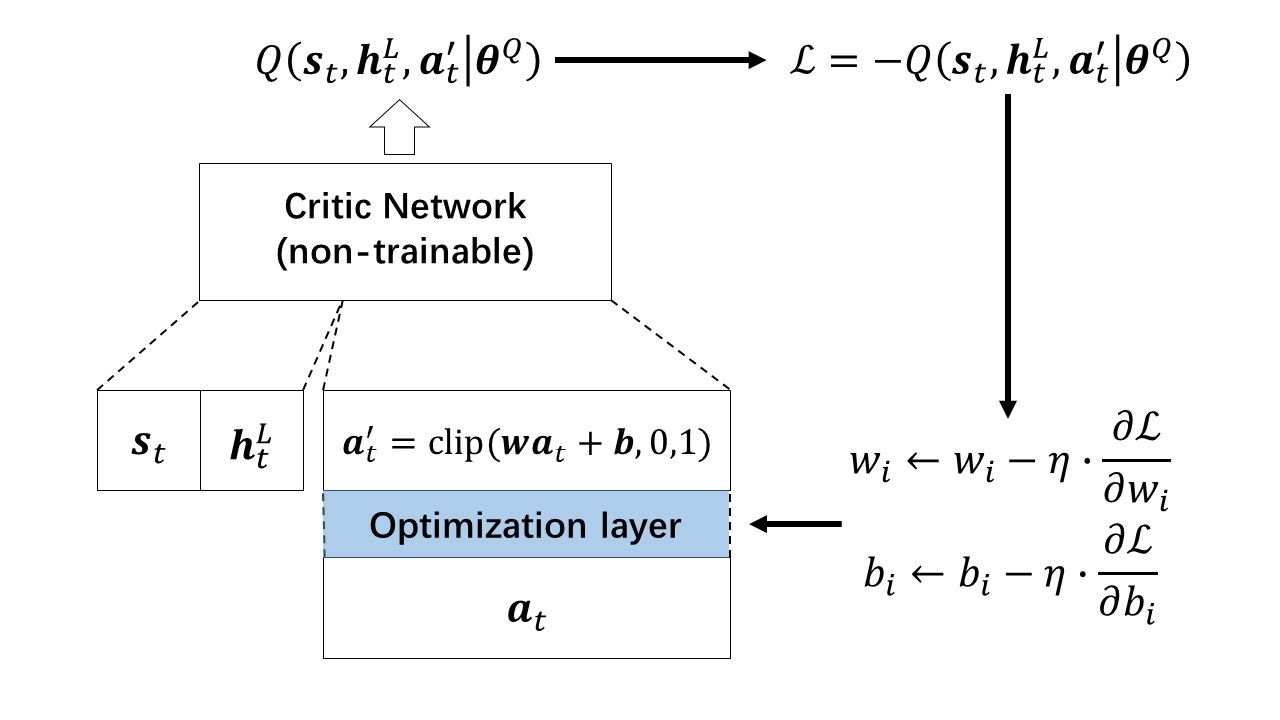}
    \caption{Illustration of the refiner module}
    \label{fig: refiner_module}
\end{figure}

The refiner module is illustrated in Figure~\ref{fig: refiner_module}. The core component of the refiner module is an optimization layer inserted between the actor’s action output and the critic network. Given the actor’s original action $\bm{a}_t = \{a_i\}_{i\in\mathcal I}$, the optimization layer applies a simple linear affine transformation followed by:
\begin{linenomath}
\begin{equation}
    \bm{a}'_t = \{a'_i\}_{i\in\mathcal I} = \big\{\text{clip}(w_ia_i + b_i, 0, 1)\big\}_{i\in \mathcal I} \label{equ: refined action}
\end{equation}
\end{linenomath}
where $\bm{w} = \{w_i\}_{i\in\mathcal I}$ and $\bm{b} = \{b_i\}_{i\in\mathcal I}$ are trainable parameters with dimensionality matching the action space. And the clipping function $\text{clip}(x,a,b) = \min(\max(x,a),b)$ ensures that the refined actions remain within valid bounds (e.g., a normalized proportion of drivers accepting discount orders between 0 and 1). This transformation allows each action dimension to be independently scaled and shifted, providing localized refinement capacity without introducing a large number of additional trainable parameters.

Importantly, the critic network used in this refinement process remains frozen (non-trainable). However, it still plays a crucial role in optimization by providing value gradients $\partial Q/\partial a'_i$ with respect to the refined action $\bm{a}'_t$. During refinement, the output of the optimization layer $\bm{a}'_t$ is fed into the critic, and the estimated Q value is used to define the ``loss'' we intend to minimize:
\begin{linenomath}
\begin{equation}
    \mathcal{L} = - Q(\bm{s}_t, \bm{h}_t^L, \bm{a}'_t|\bm{\theta}^Q)
\end{equation}
\end{linenomath}
Gradients are then backpropagated through the optimization layer to update $\bm{w}$ and $\bm{b}$, using standard gradient descent:
\begin{linenomath}
\postdisplaypenalty=0
\begin{align}
    & w_i \leftarrow w_i - \eta\frac{\partial \mathcal L}{\partial w_i} = w_i + \eta\frac{\partial Q}{\partial a_i'}\frac{\partial a_i'}{\partial w_i}, \forall i \in \mathcal I \nonumber \\
    & b_i \leftarrow b_i - \eta\frac{\partial \mathcal L}{\partial b_i} = b_i + \eta\frac{\partial Q}{\partial a_i'}\frac{\partial a_i'}{\partial b_i}, \forall i\in \mathcal I \label{equ: parameter updating of optimization layer}
\end{align}
\end{linenomath}
Since the refined action is defined as $a_i' = \text{clip}(w_i a_i^0 + b_i, 0, 1)$, its derivatives with respect to $w_i$ and $b_i$ are given by:
\begin{linenomath}
\postdisplaypenalty=0
\begin{align}
& \frac{\partial a_i'}{\partial w_i} =
\begin{cases}
a_i & \text{if } 0 < w_i a_i + b_i < 1 \\
0 & \text{otherwise}
\end{cases}, \forall i \in \mathcal I
\\
& \frac{\partial a_i'}{\partial b_i} =
\begin{cases}
1 & \text{if } 0 < w_i a_i + b_i < 1 \\
0 & \text{otherwise}
\end{cases}, \forall i \in \mathcal I
\end{align}
\end{linenomath}
These expressions indicate that gradients flow through the optimization layer only when the pre-clipped value lies strictly within the valid action bound. When the affine transformation falls outside $(0, 1)$, the clipping operation becomes saturated and the gradient vanishes, halting future-step parameter updates for that action dimension.

The refiner is able to adaptively fine-tune the actor’s decisions on a per-step basis, guided directly by the current value landscape learned by the critic. The optimization layer is re-initialized every time the refiner module is invoked during DDPG learning, typically with $w_i = 1$ and $b_i = 0, \forall i \in \mathcal I$, ensuring that its initial output exactly replicates the actor’s original action. This initialization strategy allows the module to act as an identity transformation for the first step, and then gradually adjust the action only if value improvements can be achieved through local optimization.
 
\subsection{Online action refinement}

The refiner is invoked during each decision step to improve the actor’s output through local value-guided optimization. At a given state $\bm{s}_t$ with historical latent features $\bm{h}_t^L$, the actor produces a base action $\bm{a}_t = \mu(\bm{s}_t, \bm{h}_t^L|\bm \theta^\mu)$. The refiner then initializes its optimization layer parameters to an identity mapping, i.e., $w_i = 1$ and $b_i = 0$ for all $i \in \mathcal{I}$, such that the refined action initially equals the actor’s output. A short inner-loop optimization process is then executed to iteratively adjust these parameters based on the critic’s feedback. In each refinement step, the current refined action $\bm{a}_t'$ is obtained via a forward pass through the optimization layer, followed by a critic evaluation to compute the Q-value $Q(\bm{s}_t, \bm{h}_t^L, \bm{a}_t'|\bm \theta^Q)$. Using the loss $\mathcal{L} = -Q(\bm{s}_t, \bm{h}_t^L, \bm{a}_t'|\bm \theta^Q)$, gradients with respect to $\bm{w}$ and $\bm{b}$ are computed via backpropagation, and a gradient descent update is applied to improve the refined action by Equation~\ref{equ: parameter updating of optimization layer}. This process is repeated for a fixed number of refinement steps $K_{\text{refine}}$, after which the final refined action is computed and passed to the environment for execution.

\begin{algorithm}[ht!]
	\caption{Online Action Refinement via Refiner Module}
	\label{algo: online refinement}
	\begin{algorithmic}[1]
        \Require Actor output $\bm{a}_t = \mu(\bm{s}_t, \bm{h}_t^L|\bm \theta^\mu)$; Critic network $Q(\bm{s}, \bm{a}, \bm{h}^L|\bm \theta^Q)$; refinement step count $K_{\text{refine}}$
        \Ensure Refined action $\bm{a}'_t$ to be executed in the environment
        \State Initialize optimization layer parameters $w_i \gets 1$, $b_i \gets 0$, for all $i \in \mathcal{I}$
        \For{$k \in \{1, \cdots, K_{\text{refine}}\}$}
            \State Compute refined action $\bm{a}'_t$ by Equation~\ref{equ: refined action}
            \State Evaluate Q-value $Q(\bm{s}_t, \bm{a}'_t, \bm{h}_t^L|\bm \theta^Q)$ using the current critic network
            \State Compute gradients and update parameters $\bm{w}, \bm{b}$ using Equation~\ref{equ: parameter updating of optimization layer}
        \EndFor
        \State \Return Final refined action $\bm{a}'_t$
	\end{algorithmic}
\end{algorithm}

To balance efficiency and training stability, we employ a progressive refinement schedule where the number of refinement steps increases with training progress. For example, the refinement step count can be defined as $K_{\text{refine}} = \min(ep, K_{\max})$, where $ep$ denotes the current DDPG training episode and $K_{\max}$ is a predefined upper bound. This design is motivated by the fact that in early training episodes, the critic is still under-trained and may provide unreliable value estimates. Allowing aggressive refinement based on such noisy gradients can lead to unstable updates or poor local optima. By starting with minimal refinement and gradually increasing the refinement depth as the critic becomes more accurate, the learning framework can safely integrate the refiner module and ensure that the refinement process is guided by more trustworthy value information.

This online refinement mechanism effectively enables dynamic, per-step adjustment of the policy in response to the current value landscape, leveraging the critic's guidance without modifying the underlying actor network. It provides a flexible plug-in approach that improves decision quality while maintaining the stability and generalizability of the actor.

\subsection{Policy-Improved DDPG learning}

A complete pi-DDPG algorithm for drivers' discount order acceptance setting control is illustrated in Algorithm~\ref{algo: pi-DDPG}.

\begin{algorithm}[ht!]
	\caption{pi-DDPG for Drivers' Discount Order Acceptance Setting Control}
	\label{algo: pi-DDPG}
	\begin{algorithmic}[1]
	\State Initialize critic $Q(\bm{s}, \bm{h}^L, \bm{a}|\bm{\theta}^Q)$, actor $\mu(\bm{s}, \bm{h}^L|\bm{\theta}^\mu)$, and their target networks $Q'$, $\mu'$
	\State Initialize prioritized replay buffer $\bm{R}$ and Gaussian noise $\mathcal N(0,\sigma^2)$
	\For{each episode $i \in \{1,\cdots,K\}$}
	    \State Observe initial state $\bm{s}_0$ and historical latent state $\bm{h}_0^L$
	    \State Set refinement step count $K_{\text{refine}} \gets \min(i, K_{\max})$
	    \For{each time step $t \in \{0, \cdots, T\}$}
	        \State Select base action $\bm{a}_t = \mu(\bm{s}_t, \bm{h}_t^L|\bm{\theta}^\mu)$
	        \State Obtain refined action $\bm{a}_t'$ using Algorithm~\ref{algo: online refinement} with $K_{\text{refine}}$ steps
	        \State Execute the refined action with noise for exploration $\bm{a}_t' \leftarrow \bm{a}_t' + \mathcal N(0, \sigma^2)$
            \State Observe reward $r_t$, next state $\bm{s}_{t+1}$ and $\bm{h}_{t+1}^L$
	        \State Store transition $(\bm{s}_t, \bm{h}_t^L, \bm{a}'_t, r_t, \bm{s}_{t+1}, \bm{h}_{t+1}^L)$ in prioritized buffer $\bm{R}$, ranked by $|\delta_t|$
	        \State Sample a minibatch of $N$ transitions $\{(\bm{s}_j, \bm{h}_j^L, \bm{a}_j, r_j, \bm{s}_{j+1}, \bm{h}_{j+1}^L)\}_{j=1}^N$ from $\bm{R}$ with PER
	        \State Compute TD error $\delta_j = y_j - Q(\bm{s}_j, \bm{h}_{j}^L, \bm{a}_j|\bm \theta^Q)$
	        \State Compute importance-sampling weights for the sampled mini-batch by Equation~\ref{equ: importance-sampling weights}
            \State Update sampled transition $j$'s rank according to its absolute TD-error $|\delta_j|$
	        \State Update critic $Q(\bm{s}, \bm{h}^L, \bm{a}|\bm{\theta}^Q)$ by minimizing the weighted loss by Equation~\ref{equ: weighted loss}
	        \State Update actor $\mu(\bm{s}, \bm{h}^L|\bm{\theta}^\mu)$ using the policy gradient by Equation~\ref{equ: policy gradient}
	        \State Soft update target networks' weights $\bm{\theta}^{Q'}, \bm{\theta}^{\mu'}$ by Equation~\ref{equ: soft update}
	    \EndFor
        \State Decline $\sigma$ of the Gaussian noise
	\EndFor
	\end{algorithmic}
\end{algorithm}

%% file: data/section6_experiment.tex
\section{Simulation and Experiments} \label{sec 6}

To demonstrate the effectiveness of the proposed pi-DDPG, a simulator was constructed as an interactable virtual learning environment for deep reinforcement learning agents. Real-world ride-hailing operational data was processed to calibrate key parameters to depict the spatiotemporal distributions of supply and demand within the simulation. We compare the performance of pi-DDPG against a benchmark DDPG algorithm that does not include the refiner module, under a variety of experimental scenarios. The comparison focuses on learning efficiency, with particular attention to the episode rewards achieved during the early stages of training.

\subsection{Simulator}
\begin{figure}[htbp]
    \centering
    \includegraphics[width=0.75\linewidth]{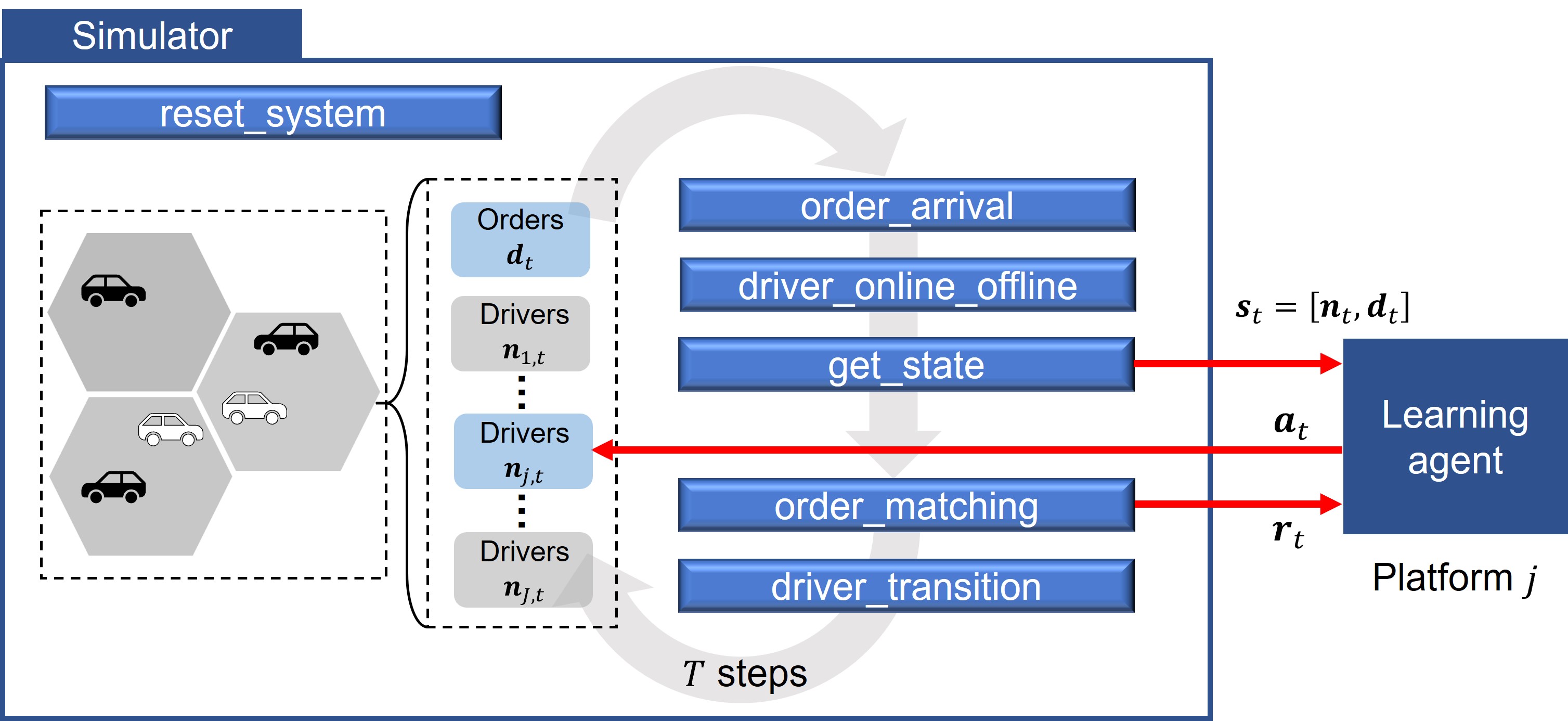}
    \caption{Simulator structure and its core functions}
    \label{fig: simulator structure}
\end{figure}

A customized simulator is developed to validate the effectiveness of the proposed pi-DDPG. It replicates multi-platform ride-hailing operations within an urban setting and enables the learning agent to interact with a dynamic, stochastic environment through sequential decision-making over multiple time intervals. The overall architecture and workflow of the simulator are depicted in Figure~\ref{fig: simulator structure}. Each learning episode begins with a system reset, which initializes the environment state. The environment consists of a spatial network divided into multiple hexagonal grids, with drivers and orders distributed across locations. At each time step $t$, the simulator proceeds through a series of functions:
\begin{itemize}
    \item [1)] \texttt{order\_arrival}: Newly-emerging requests are generated according to a Poisson process. These orders are categorized by type and location.
    \item [2)] \texttt{driver\_online\_offline}: For each individual platform, newly online drivers are generated according to a Poisson process, and driver offline behavior is modeled as a Bernoulli process.
    \item [3)] \texttt{get\_state}: The system sends the current observable state to the learning agent representing individual platform $j$, including the spatial distribution of online drivers affiliated with individual platform $j$ and global order distribution.
    \item [4)] \texttt{order\_matching}: Based on the actions from all individual platforms, the integrator performs centralized order matching. The learning agent receives an immediate reward based on the matching result.
    \item [5)] \texttt{driver\_transition}: After order completion, matched drivers are moved to their respective destination grids, and unmatched drivers remain in place.
\end{itemize}
This process is repeated for $T$ time steps, corresponding to one full episode. Throughout training, the simulator supports learning-by-interaction, allowing the agent to explore, refine, and evaluate its control policy under competitive multi-platform conditions. The simulator was developed using Python 3.11 and the \href{https://github.com/openai/gym}{gym 0.26.2 package}. For this study, both the simulation and neural network training were conducted on a personal computer running Windows 11, equipped with an Intel i5-13600KF CPU (3.9 GHz), 32 GB of DDR4 RAM (3200 MHz), and an RTX 4070 GPU (12 GB of VRAM).

\subsection{Data processing}
\label{subsec: data processing}

This study utilizes real-world passenger demand and driver supply data obtained from a major integrated ride-hailing platform operating in Beijing, covering the period from April 1 to June 30, 2024. Due to the sparsity of data in suburban regions, the analysis focuses exclusively on the central urban area. This area is spatially partitioned into 37 grids using Uber's H3 index system at resolution level 6, as shown in Figure~\ref{fig: hexagonal grid encoding}. 
The demand data comprises the life-cycle information of each request, including order ID, generation time, selected platforms, order type, order and destination coordinates, assignment/cancellation information, and the actual drop-off time. To be more specific, a detailed classification of order types is provided in Table~\ref{tab: order type}. The supply dataset tracks the operational status drivers from different individual platforms. 
Each record details the time a driver spends in various operational states within a five-minute interval, including waiting, picking-up, delivering the matched services, and actively listening for Discount Express or Standard Express requests. It also records the driver’s location at the end of each interval. In addition, the dataset includes drivers' static attributes, such as driver ID, affiliated individual platform, and designated service class (e.g., Express or Premium).

\begin{table}[htbp!]
\centering
\caption{Order type classification based on service selection}
\label{tab: order type}
\begin{tabular}{cccc}
\hline
 Order type & Discount Express service & Standard Express service & Premium service \\
\hline
 1 & \checkmark &  &  \\
 2 &  & \checkmark &  \\
 3 & \checkmark & \checkmark &  \\
 4 &  &  & \checkmark \\
 5 & \checkmark &  & \checkmark \\
 6 &  & \checkmark & \checkmark \\
 7 & \checkmark & \checkmark & \checkmark \\
\hline
\end{tabular}
\end{table}

Based on empirical data analysis, Premium drivers seldom serve Discount Express requests, and Express drivers can not provide Premium services. Moreover, Type 5, Type 6,and Type 7 orders -- where passengers simultaneously select both Express and Premium service options -- account for only a small fraction of the demand pool. Therefore, to simplify the simulation, this simulator excludes Premium drivers and passengers requesting Premium services, and neglects the potential influence of Premium services on Express service. 
Accordingly, the simulation and numerical experiments consider only Type 1, Type 2, and Type 3 orders, and restrict the driver pool to  Express drivers.

\begin{figure}[htbp]
	\begin{minipage}{0.48\textwidth}
		\centering
		\includegraphics[width=\linewidth]{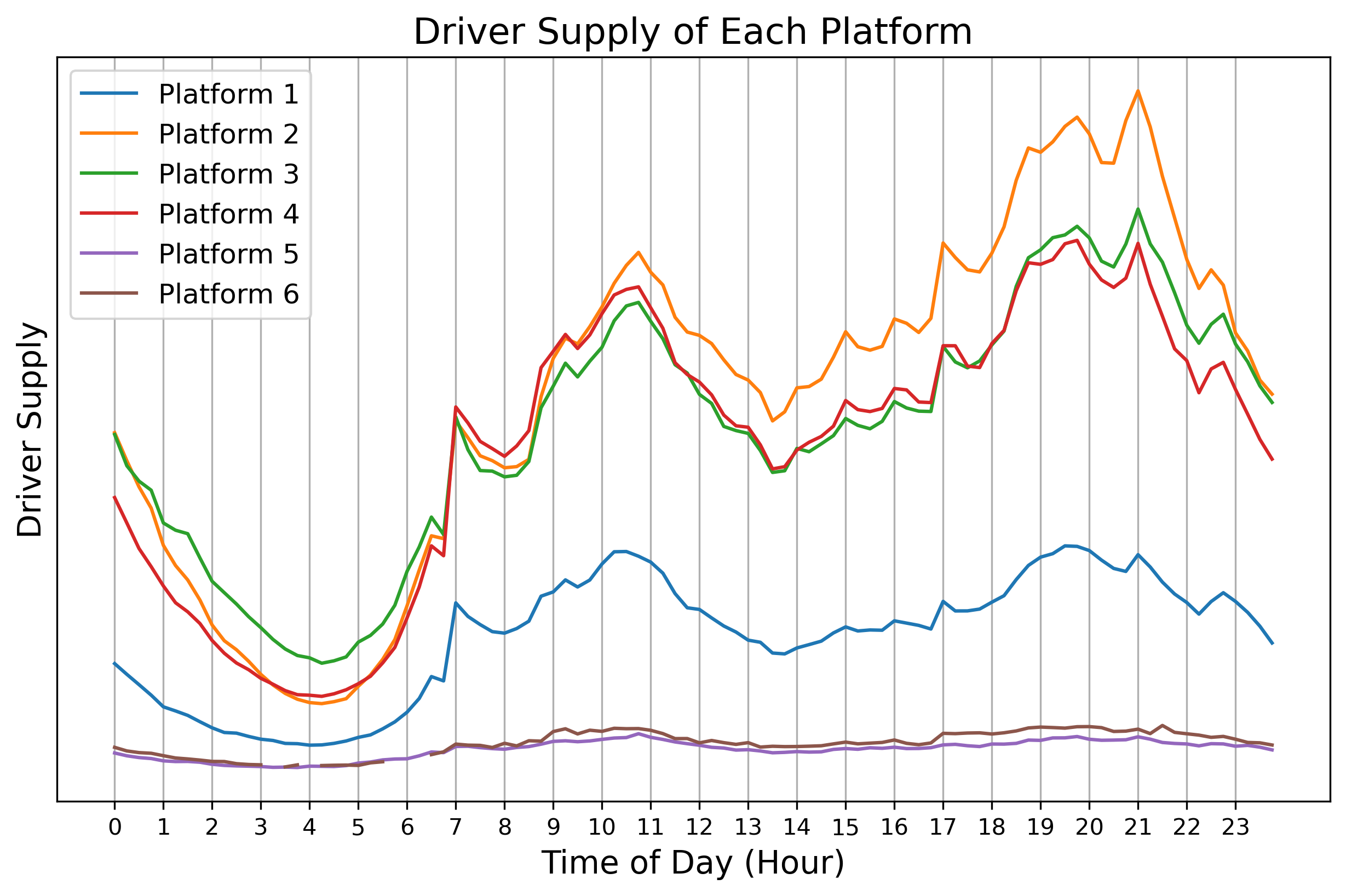}
	\end{minipage}
	\begin{minipage}{0.48\textwidth}
		\centering
		\includegraphics[width=\linewidth]{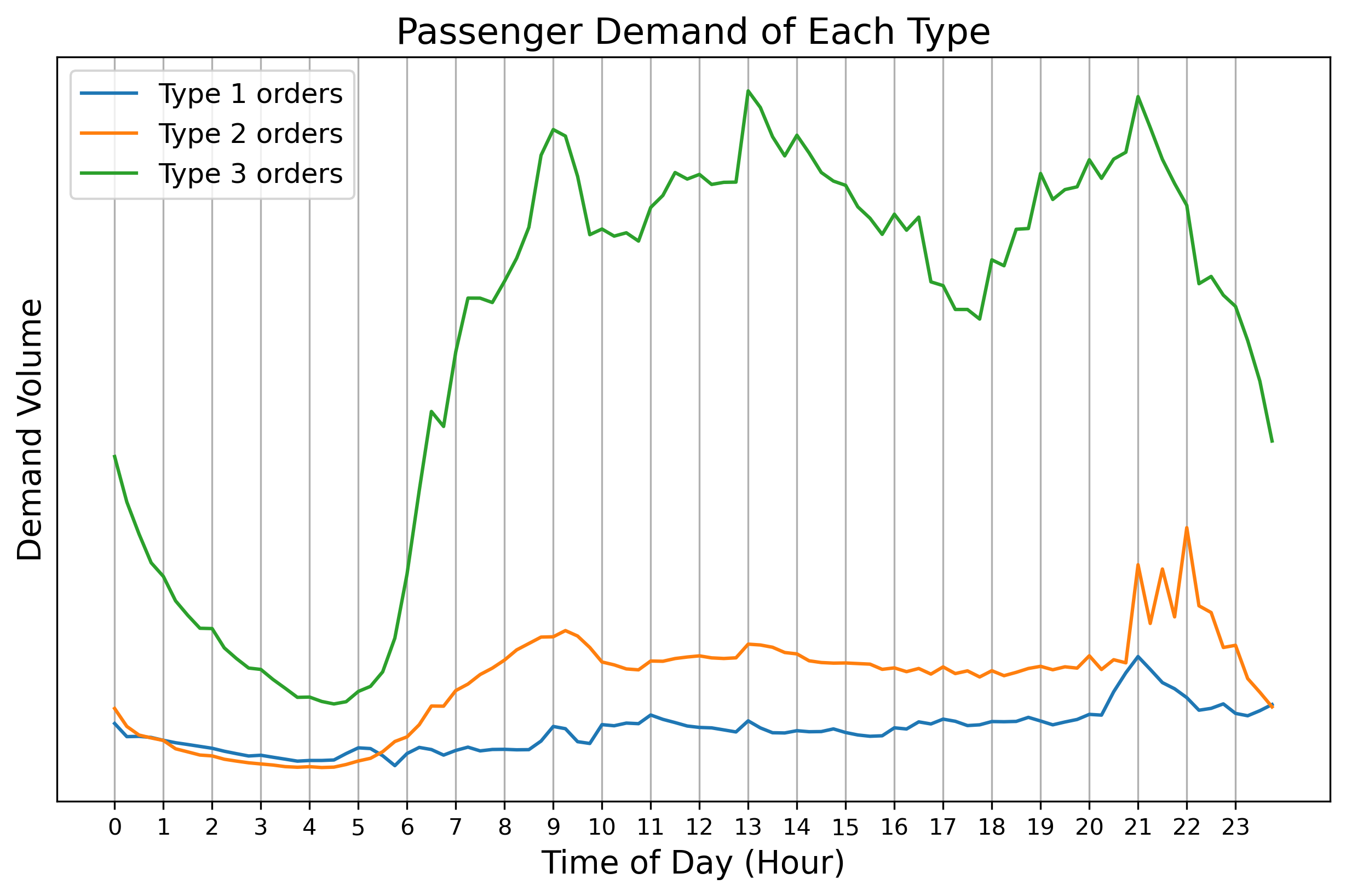}
	\end{minipage}
    \caption{Trend of driver supply and passenger demand over a day} 
	\label{fig: hourly trend of supply and demand}
\end{figure}

\begin{figure}[htbp]
    \centering
    \includegraphics[width=0.5\linewidth]{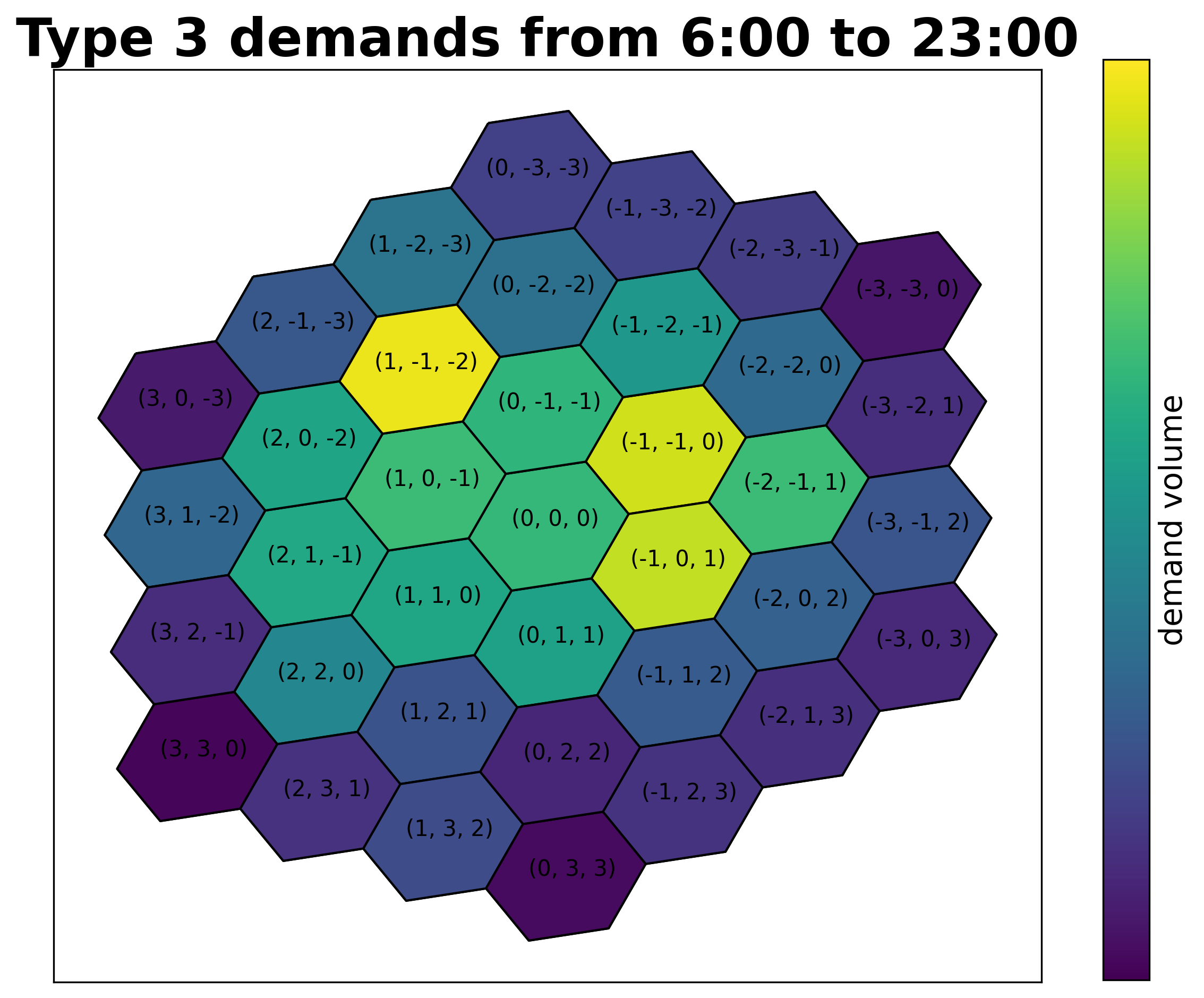}
    \caption{Spatial distribution of passenger demands}
    \label{fig: spatial distribution of demand}
\end{figure}

The inputs of simulation include temporal dynamics of order arrivals, the spatial distributions of order origins and destinations, and the online/offline transition rates of drivers. Both order demand and driver supply exhibit strong daily periodicity. To capture these dynamics, the time horizon is discretized into 15-minute time intervals. Let $\overline d _{i,j,t}$ denote the average number of order arrivals of order type $(j)$ in grid $(i)$ during time interval $(t)$. The arrival of new orders is modeled as a heterogeneous Poisson process, with rate parameter, with parameter $\lambda^{d}(t) = \overline d _{i,j,t}$. These values are calibrated using historical data on Type 1-3 requests based on their origin locations and request generation times, as illustrated in Figure~\ref{fig: hourly trend of supply and demand} and Figure~\ref{fig: spatial distribution of demand}. For simplification, we assume that each order selects all available individual platforms, implying a shared order pool across all platforms.
Similarly, the average number of online drivers for each individual platform $(m)$ within time intervals $(t)$, denoted by $\overline n_{m,t}$, is used to model driver supply. The arrival of newly-online drivers for each individual platform $m$ also follows a heterogeneous Poisson process. The corresponding parameter $\lambda^{s}(t)$ is determined by the expected net increase in the number of drivers from interval $t$ to $t+1$:
\begin{linenomath}
\begin{equation}
    \lambda^{s}(t) = \begin{cases}
       0, & \text{if }\overline n_{m,t+1} \le \overline n_{m,t}\\
       \overline n_{m,t+1} - \overline n_{m,t}, & \text{otherwise}
    \end{cases}
\label{equ: calibrate driver online}
\end{equation}
\end{linenomath}
Newly-online drivers are assumed to be uniformly distributed across all grids. The driver's offline behavior is modeled as an independent, memoryless binomial process. At the end of each time interval, a driver has a probability $p(t)$ of going offline, defined as:
\begin{linenomath}
\begin{equation}
    p(t) = \begin{cases}
       0, & \text{if }\overline n_{m,t+1} \ge \overline n_{m,t}\\
       (\overline n_{m,t} - \overline n_{m,t+1})/\overline n_{m,t}, & \text{otherwise}
    \end{cases}
\label{equ: calibrate driver offline}
\end{equation}
\end{linenomath}
Notably, Equations~(\ref{equ: calibrate driver online}) and Equation~(\ref{equ: calibrate driver offline}) serve only to calibrate the simulator environment. The RL agent’s state input directly consists of the aggregate number of active drivers, and the learning algorithm does not relies on explicitly modeling drivers' online/offline process.

\subsection{Experiment settings}

Our proposed pi-DDPG algorithm is evaluated against a \textbf{baseline DDPG} (Algorithm~\ref{algo: DDPG}), \textcolor{red}{which is} introduced in Section~\ref{sec 4}. Both algorithms incorporate spatiotemporal memory by encoding environmental inputs into a three-dimensional matrix and processing them via a ConvLSTM network. They also share a prioritized experience replay buffer to improve sampling efficiency during training. The only difference between the two approaches lies in the inclusion of the refiner module in pi-DDPG.

Layers in the deep neural networks are initialized using He Normal, Glorot, or Random Uniform initialization schemes, depending on the activation functions used in each layer. Specifically:
\begin{itemize}
    \item For layers using ReLU activations, we apply He Normal Initialization (\citealp{He_2015_ICCV}), where each weight $w$ is drawn from a zero-mean Gaussian distribution with variance: $w\sim \mathcal N(0, \frac{2}{n_{\text{in}}})$, where $n_{\text{in}}$ is the number of input units to the layer.
    \item For the actor network's output layer with a tanh activation, we adopt the Glorot (Xavier) Initialization (\citealp{pmlr-v9-glorot10a}), which draws weights from a zero-mean Gaussian distribution with variance: $w\sim \mathcal N(0, \frac{2}{n_{\text{in}} + n_{\text{out}}})$, where $n_{\text{out}}$ denotes the number of output units.
    \item For the critic network's output layer with a linear activation, we use a Random Uniform Initialization within a bounded range: $w\sim \mathcal{U}(-0.03, 0.03)$, to prevent extreme TD errors and large gradient updates in the early training, thereby stabilizing the Q value estimation process.
\end{itemize}
All hidden layer biases are initialized to zero. During training, Gaussian noise sampled from $\mathcal{N}(0, \sigma^2)$ is added to the actor’s output to encourage exploration, where $\sigma$ is initialized at 0.1 and linearly decays to zero as the number of training episodes increases. The learning rate of the actor network is set to $1\times 10^{-4}$, and the critic's learning rate is $3\times 10^{-4}$. Every time we train the actor network and the critic network, we sample a minibatch of $N = \min(|\bm R|, 128)$ transitions from the replay buffer $\bm R$. The softupdate ratio $\tau$ is $0.005$. The maximum number of online refinement steps in the refiner module is set to $K_{\max} = 10$, with the step length $\eta = 0.1$.

The simulation environment aims to closely replicate the operational characteristics of a third-party ride-hailing integrator in Beijing with the five biggest platforms. The virtual environment consists of 37 spatial grids (Figure~\ref{fig: hexagonal grid encoding}), and 3 types of passenger orders (Figure~\ref{fig: hourly trend of supply and demand}). Driver online rates and offline probabilities for each individual platform, the arrival rates of various order types on the integrator, and the driver spatial transition probabilities (O-D pair distribution of orders) are all estimated from real-world operational data. Order service rewards are determined by the spatial distance between the origin and destination grids. For example, if the origin and destination are adjacent grids, the normalized reward for a Standard Express order is set to 1. All discount orders are uniformly assigned a reward discount ratio of 0.7. Each simulation episode contains 96 intervals with equal length of 15-minute throughout a single day. The episode reward, which serves as the primary performance metric, is defined as the total revenue of order fulfillment obtained by an individual platform within one episode.

The centralized order matching mechanism of the integrator, which operates across multiple individual platforms, is defined as follows: Type 1 orders (only request Discount Express service) can only be assigned to drivers who enable the discount order acceptance setting. Type 2 orders (only request Standard Express service) will first be matched with drivers who do not accept Discount Express orders, and any remaining unmatched orders are then assigned to drivers who accept discount orders. Type 3 orders (request Discount Express or Standard Express services) are matched with the remaining available drivers, with higher priority given to those who do not accept discount orders. Notably, drivers from different individual platforms with the same order acceptance settings are treated equally during the matching process. The matching procedure also accounts for spatial distribution: orders are first matched with drivers located in the same grid; if no eligible drivers are available, the search expands to adjacent grids.

To model competition among individual platforms, we assume that other individual platforms adopt a simple control strategy in which a fixed proportion of their drivers accept discount orders. According to the statistical analysis of drivers' operations data, the fixed acceptance ratio is set as 0.3, which reflects the current average proportion of ride-hailing drivers accepting discount orders in practice.
\subsection{Experiment results}

\subsubsection{Sensitivity analysis}

\noindent (a) Sensitivity analysis of the actor network's learning rate

\begin{figure}[htbp]
    \centering
    \includegraphics[width=0.9\linewidth]{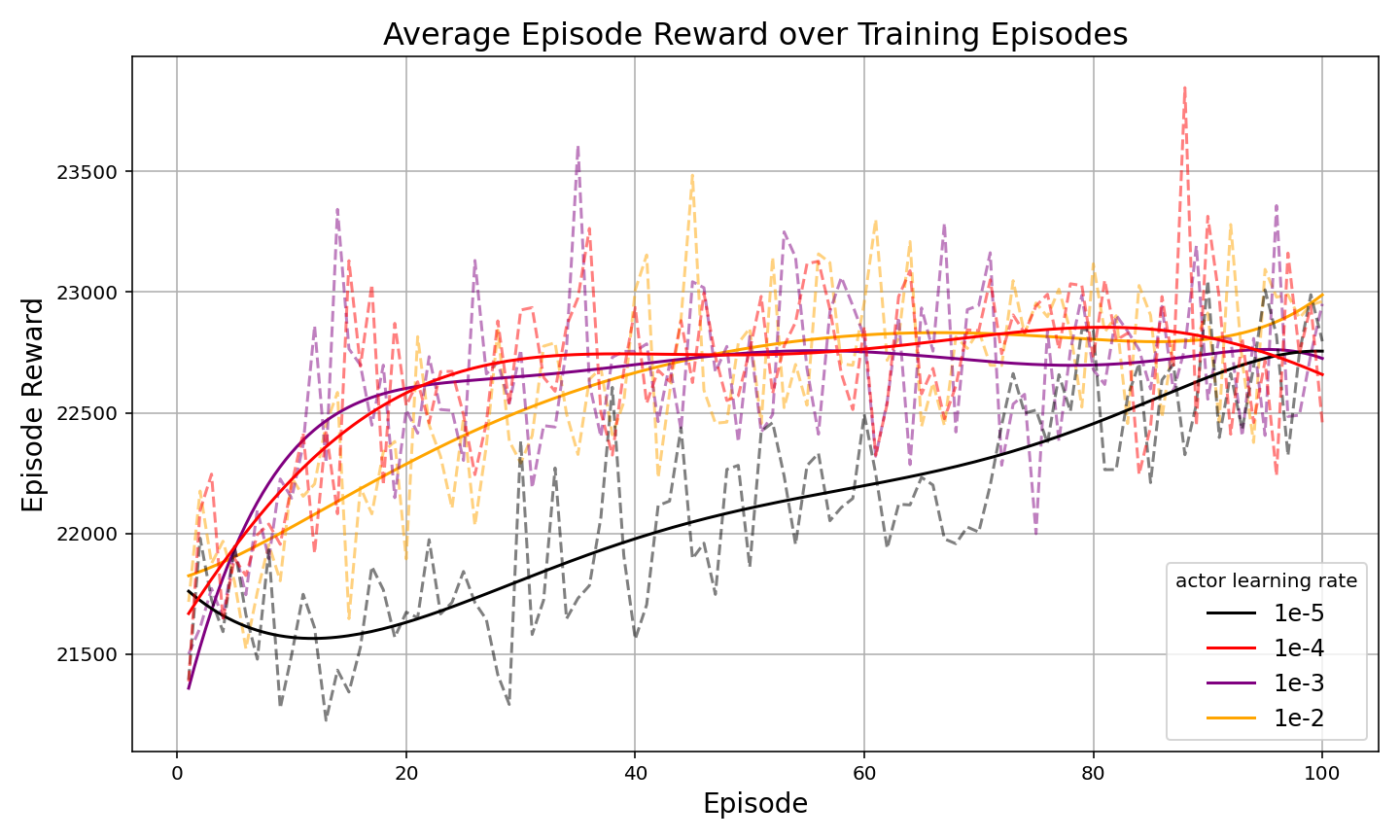}
    \caption{Comparison of training performance of DDPG under different actor learning rates}
    \label{fig: actor_lr_comparison}
\end{figure}

Figure~\ref{fig: actor_lr_comparison} presents the results of a sensitivity analysis on the learning rate of the actor network, conducted within the baseline DDPG framework. We evaluate four candidate values ranging from $10^{-2}$ to $10^{-5}$, each tested over 100 training episodes and averaged across 20 independent runs. The resulting average episode reward trajectories reveal notable differences in both learning efficiency and early-stage performance.

Among all configurations, the learning rate of $10^{-4}$ (red curve) achieves the best overall performance. It exhibits rapid early-stage convergence, reaching a high reward level within the first 20 episodes, and continues to improve steadily throughout training. In contrast, a larger learning rate of $10^{-2}$ (orange curve) performs poorly in the early training phase, while $10^{-3}$ (purple curve) converges faster initially but yields slightly lower average rewards in the later stages compared to $10^{-4}$. Meanwhile, the smallest learning rate of $10^{-5}$ (black curve) results in markedly slower convergence and limited reward growth, suggesting underfitting and insufficient policy updates.

Based on this analysis, we set the actor learning rate to $10^{-4}$ in all experiments, striking a balance between stability and learning efficiency.

\noindent (b) Sensitivity analysis of the exploration noise strength

\begin{figure}[htbp]
    \centering
    \includegraphics[width=0.9\linewidth]{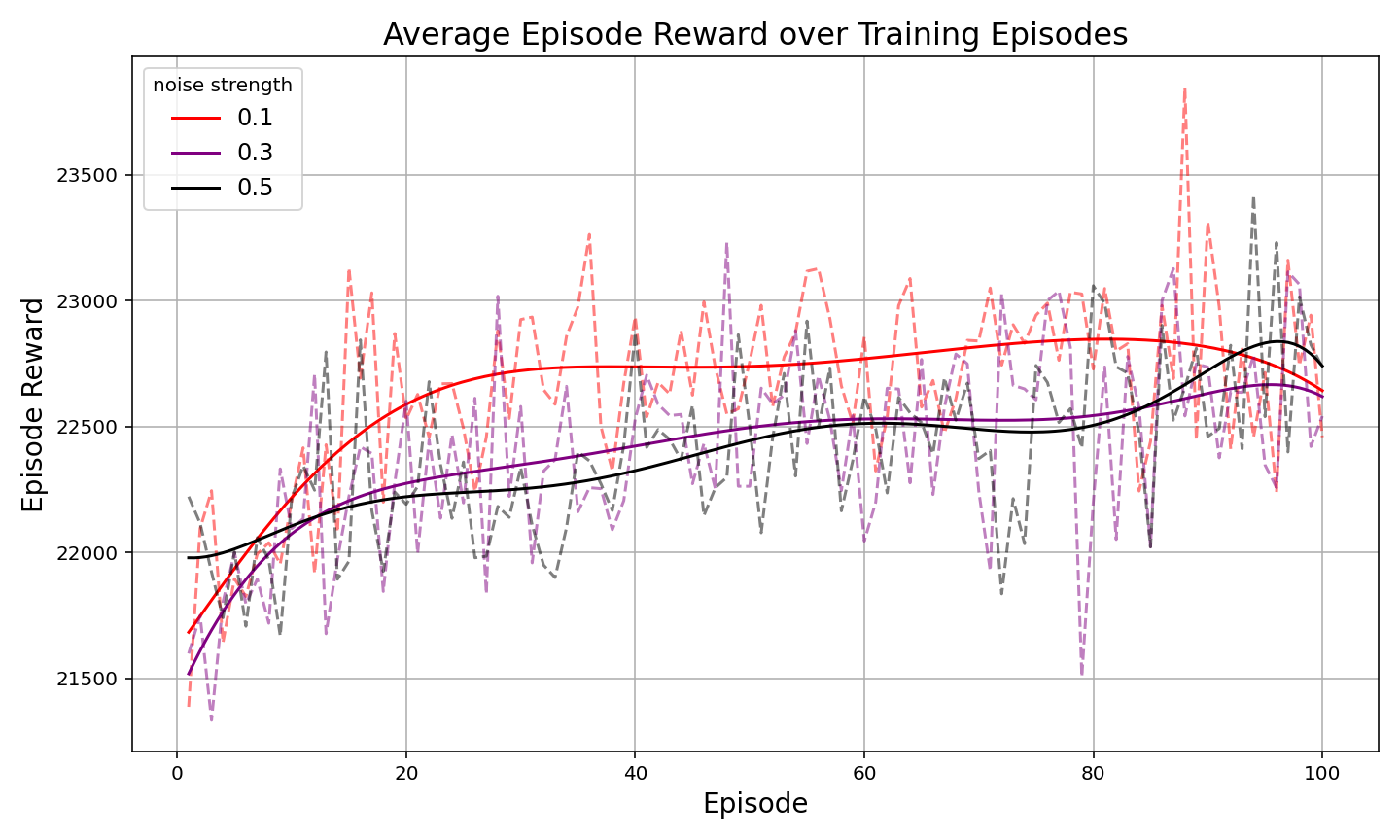}
    \caption{Comparison of training performance of DDPG under different exploration noise strengths}
    \label{fig: noise_strength_comparison}
\end{figure}

Figure~\ref{fig: noise_strength_comparison} presents the results of a sensitivity analysis on the exploration noise strength $\sigma$, conducted within the DDPG framework. We evaluate three candidate values $\{0.1, 0.3, 0.5\}$, each tested over 100 training episodes and averaged across 20 independent runs. The resulting average episode reward trajectories reveal notable differences in the algorithm's early-stage performance.

Among the three configurations, a noise strength of $\sigma = 0.1$ (red curve) achieves the most favorable performance. It converges rapidly within the first 20 episodes and maintains consistently high rewards throughout training. In contrast, larger noise strengths such as $\sigma = 0.3$ (purple curve) and $\sigma = 0.5$ (black curve) lead to poor early-stage performance. Excessive exploration at these levels increases the risk of irrational actions, which may result in reduced order match rates and operational efficiency, especially in a competitive ride-hailing market. Despite these differences in early-phase performance, all three configurations eventually converge to similar episode reward levels by the end of training.

Based on this analysis, we set the exploration noise strength $\sigma$ to $0.1$ in all experiments, as it offers a balanced trade-off between adequate exploration and early-stage learning cost.

\noindent (c) Sensitivity analysis of the maximum online refinement steps

\begin{figure}[htbp]
    \centering
    \includegraphics[width=0.9\linewidth]{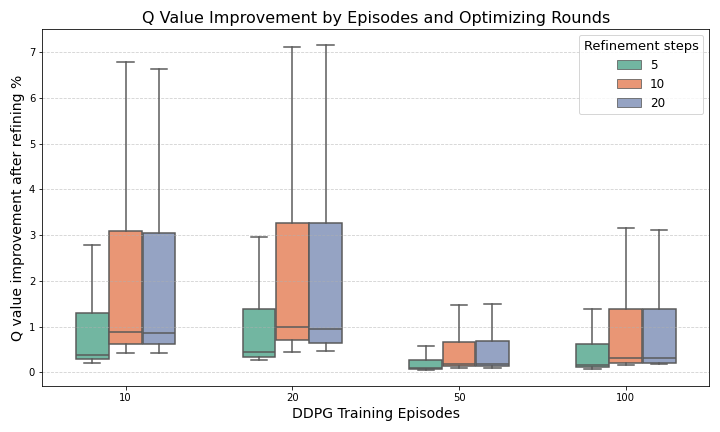}
    \caption{Q value improvement percentage with different refinement steps at different learning episodes}
    \label{fig: refinement_step_comparison}
\end{figure}

Figure~\ref{fig: refinement_step_comparison} presents the sensitivity analysis of the maximum refinement step $K_{\max}$ in the refiner module. We evaluate three configurations: 5, 10, and 20 refinement steps, and compare the critic-evaluated Q-values before and after refinement. The percentage improvement in Q-values reflects the refiner’s ability to locally enhance policy quality given the current critic under different DDPG training phases.

As shown in the figure, refinement leads to clear Q-value improvements across all episode stages, particularly in the early training phase (e.g., episode 10 and 20), where the actor policy is less mature and benefits more from local correction. The improvement magnitude grows with the number of refinement steps, but with diminishing returns: the performance gap between 10 and 20 steps is noticeably smaller than the gap between 5 and 10. From the boxplot distribution, we observe that the median Q-value improvement is generally within 0-2\%. A small number of outliers reach up to 6–8\% improvement, highlighting the potential of the refiner to recover suboptimal actions in certain cases.

Based on this trade-off between performance gain and computational cost, we set the maximum number of refinement steps $K_{\max}=10$ in our experiments. This configuration achieves substantial improvement in Q-value estimates while maintaining computational efficiency during online action refinement.

\subsubsection{Training results}

\begin{figure}[htbp]
    \centering
    \includegraphics[width=\linewidth]{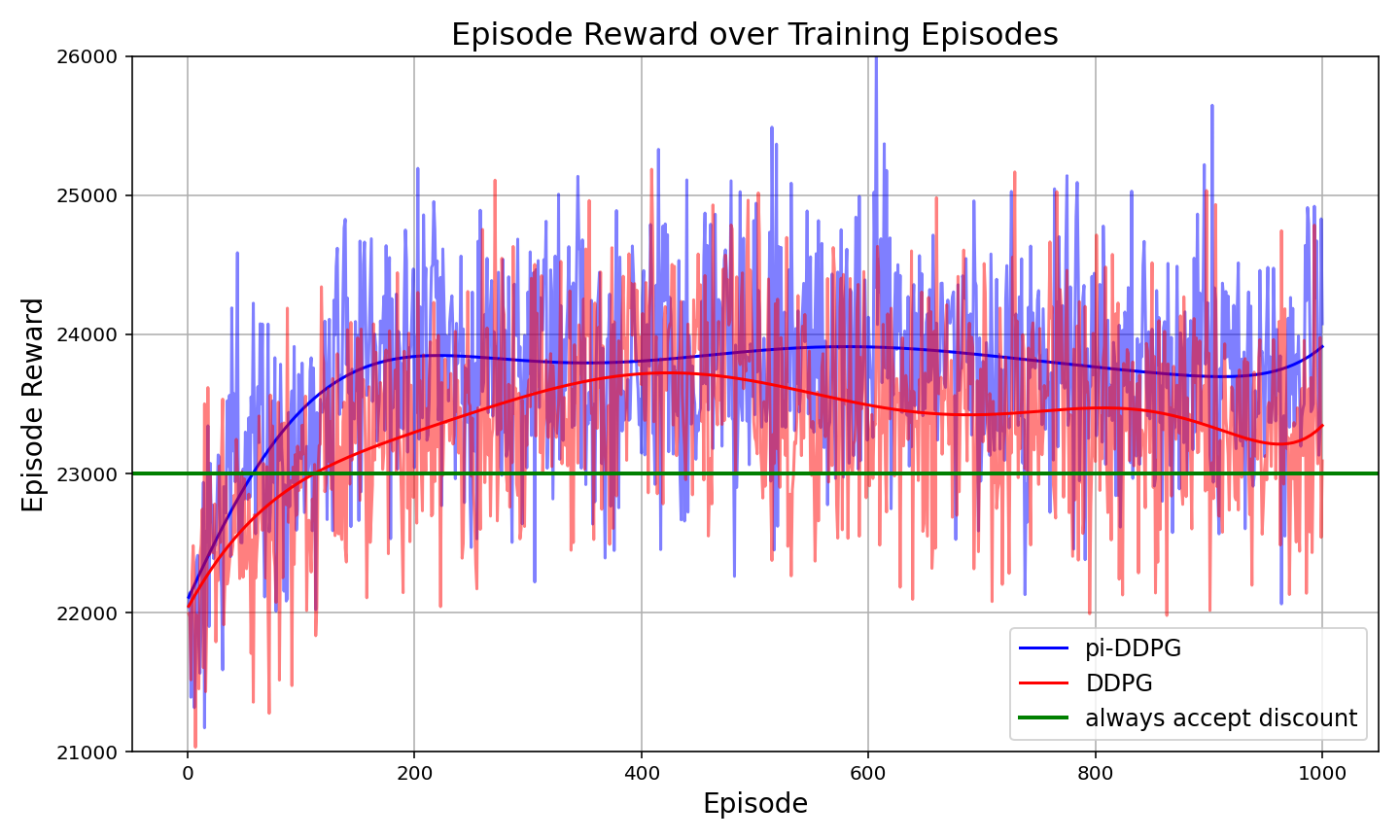}
    \caption{Comparison of training performance of DDPG and pi-DDPG for Platform 1}
    \label{fig: training results}
\end{figure}

Figure~\ref{fig: training results} illustrates the evolution of episode rewards over training episodes for the proposed pi-DDPG and the baseline DDPG. They are compared with a rule-based benchmark that always sets the discount acceptance rate to 100\% (among all constant discount acceptance rates in the range $[0,1]$, the 100\% setting yields the highest average episode reward across repeated simulations). Overall, both learning-based methods significantly outperform the fixed-policy benchmark, demonstrating the effectiveness of adaptive discount control. In the early training phase (approximately the first 400 episodes), the pi-DDPG framework exhibits a faster increase in total reward compared to the baseline DDPG, indicating the effectiveness of the online action refinement. As training progresses, both algorithms show continued improvement in episode rewards, but pi-DDPG maintains a consistently higher reward trajectory. Around Episode~400, the performance of both algorithms begins to converge. While the baseline DDPG experiences noticeable fluctuations and even reward degradation beyond this point, the pi-DDPG framework achieves a stable and sustained performance level. After pi-DDPG converges to stable performance, the green line representing the ``always accept discount'' policy remains consistently below the smoothed learning-based curve, reinforcing the importance of learning a context-aware, state-dependent discount strategy rather than applying a fixed heuristic.

\begin{figure}[htbp]
    \centering
    \begin{minipage}{0.48\textwidth}
        \centering
        \includegraphics[width=\linewidth]{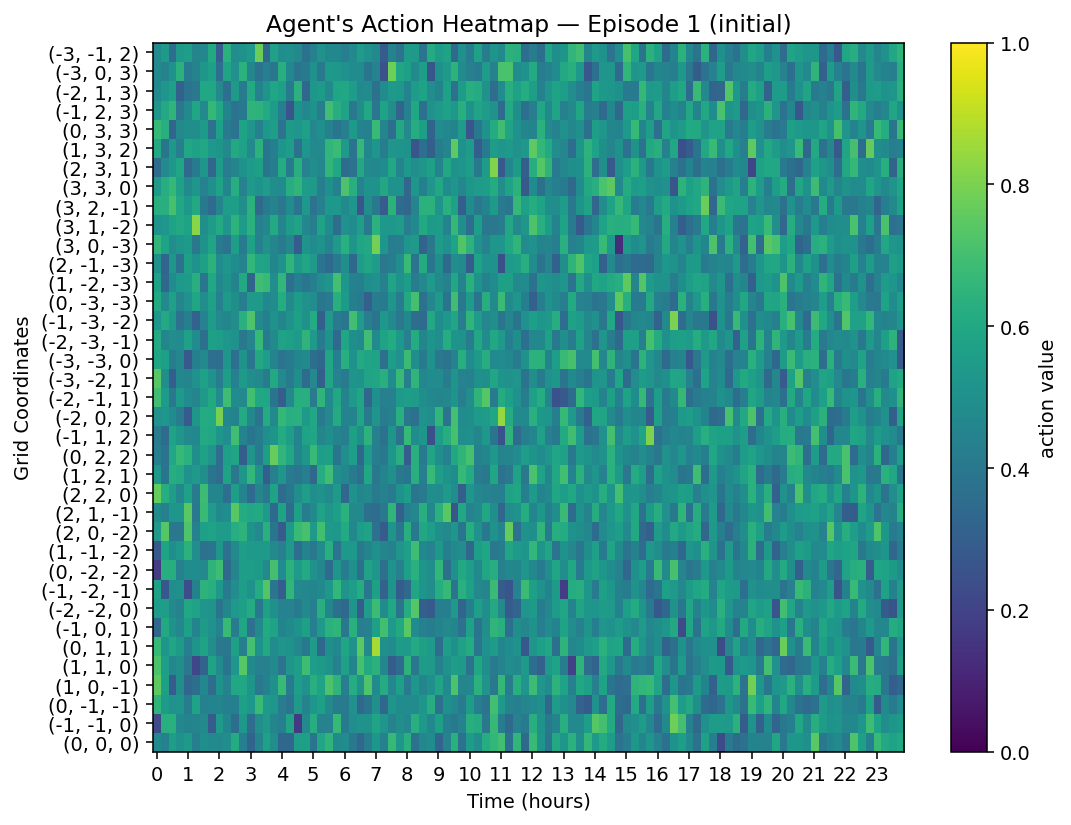}
        {\small \textbf{Episode} 1}
    \end{minipage}
    \hfill
    \begin{minipage}{0.48\textwidth}
        \centering
        \includegraphics[width=\linewidth]{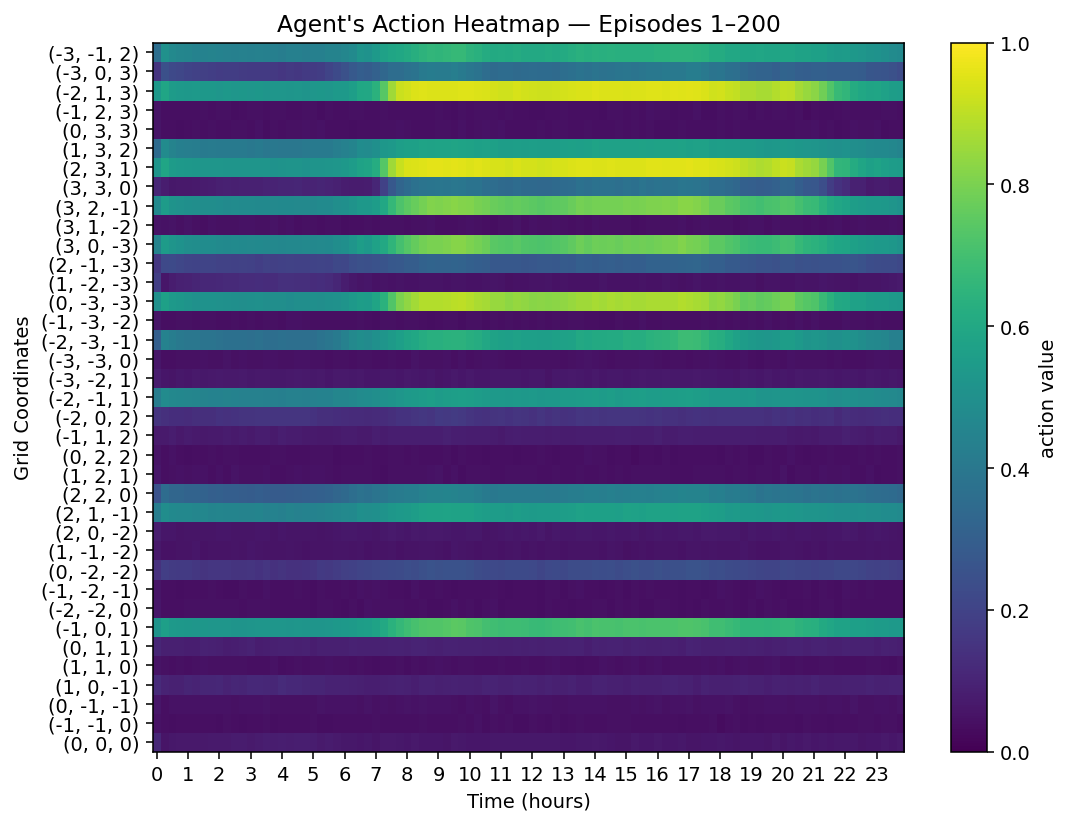}
        {\small \textbf{Episode} 1-200}
    \end{minipage}

    \vspace{1em}

    \begin{minipage}{0.48\textwidth}
        \centering
        \includegraphics[width=\linewidth]{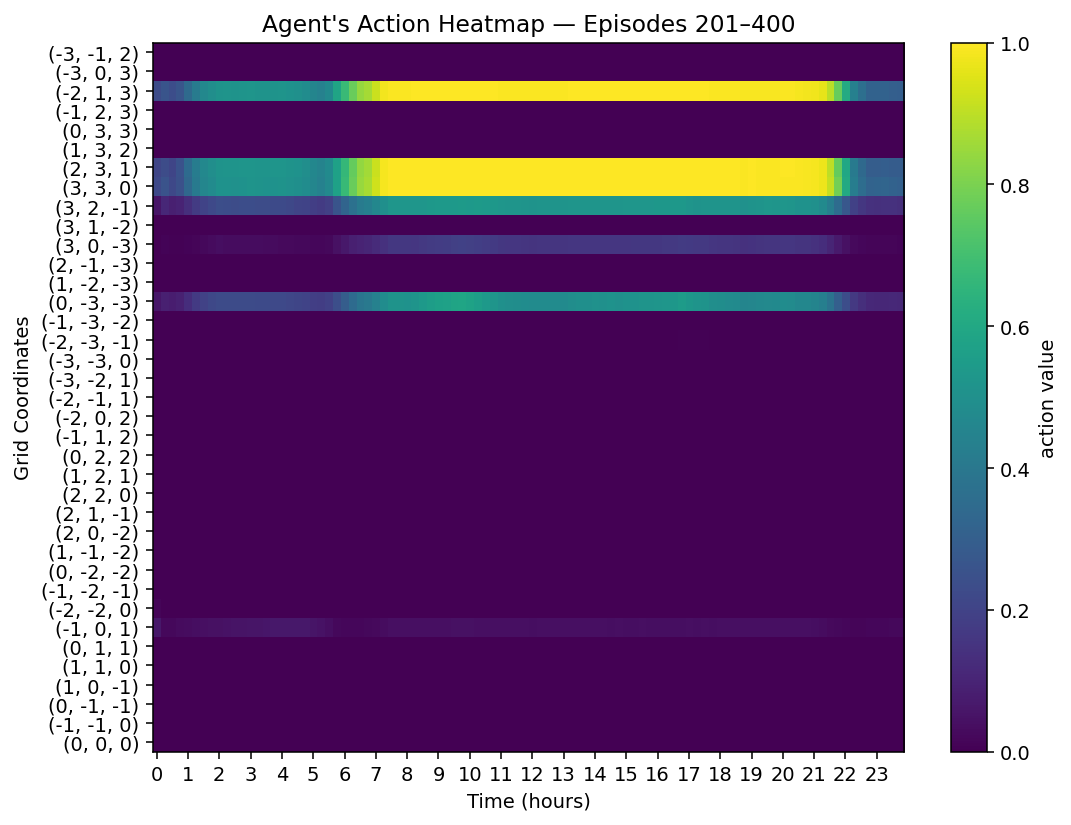}
        {\small \textbf{Episode} 201-400}
    \end{minipage}
    \hfill
    \begin{minipage}{0.48\textwidth}
        \centering
        \includegraphics[width=\linewidth]{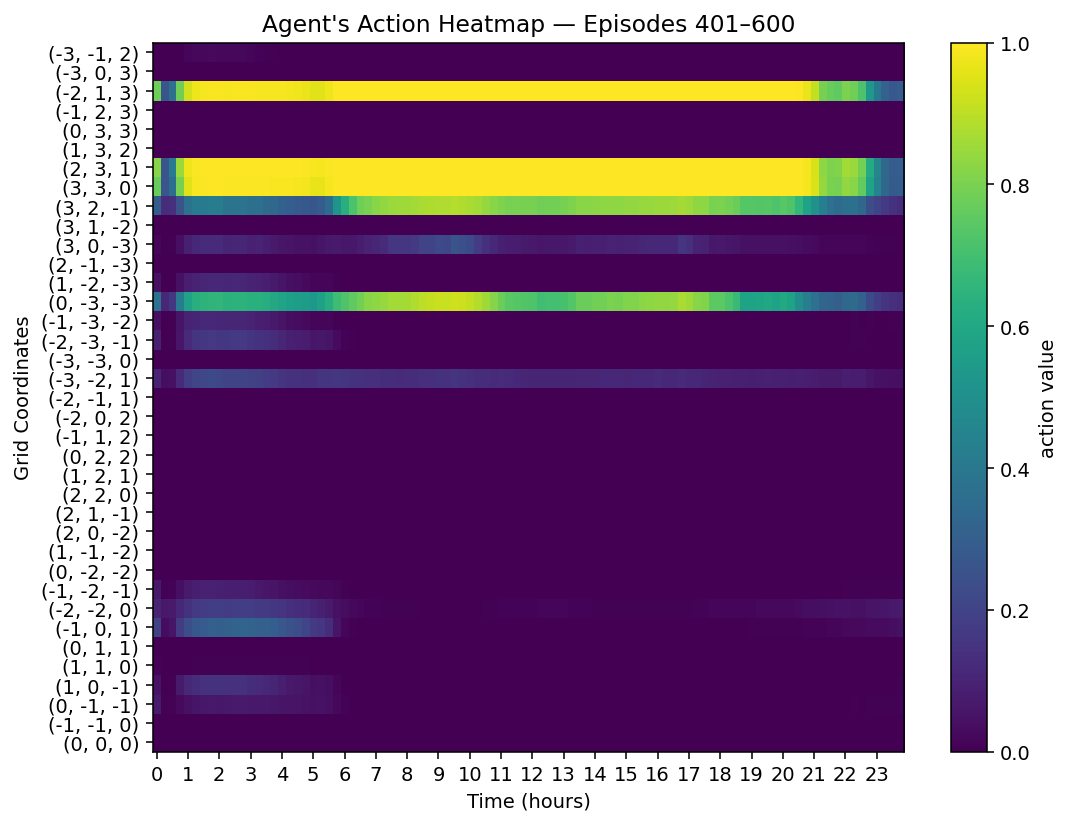}
        {\small \textbf{Episode} 401-600}
    \end{minipage}

    \vspace{1em}

    \begin{minipage}{0.48\textwidth}
        \centering
        \includegraphics[width=\linewidth]{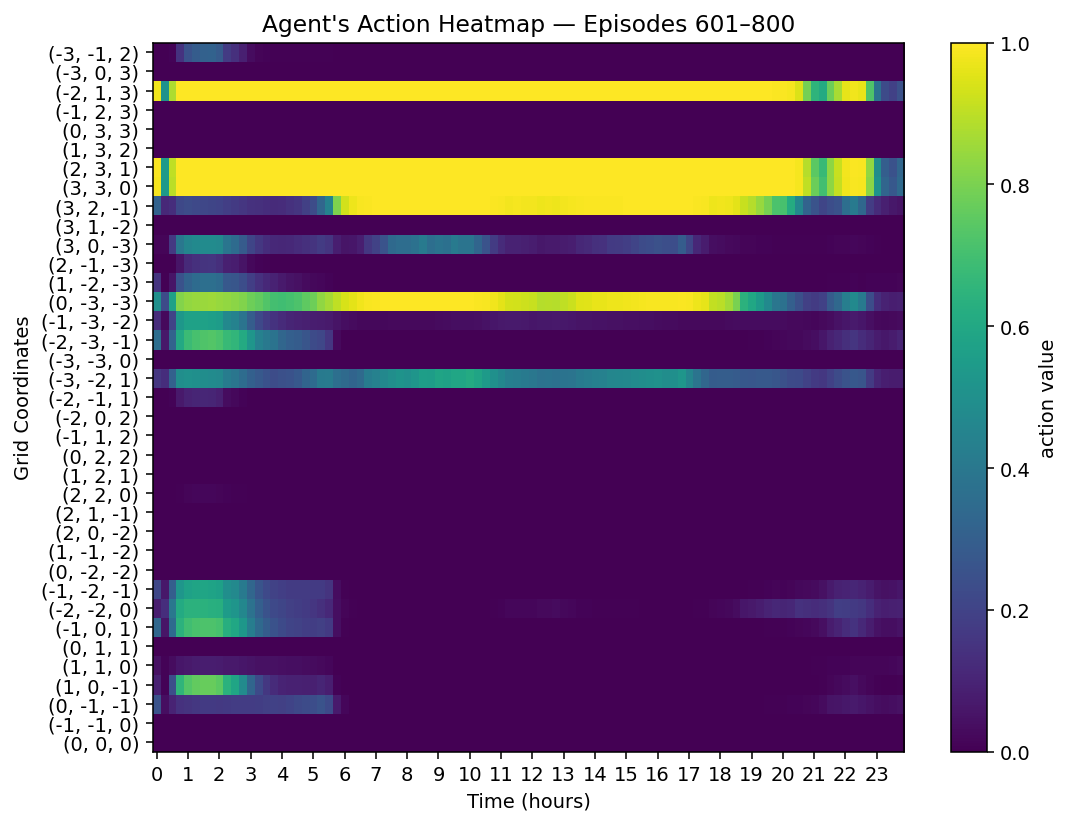}
        {\small \textbf{Episode} 601-800}
    \end{minipage}
    \hfill
    \begin{minipage}{0.48\textwidth}
        \centering
        \includegraphics[width=\linewidth]{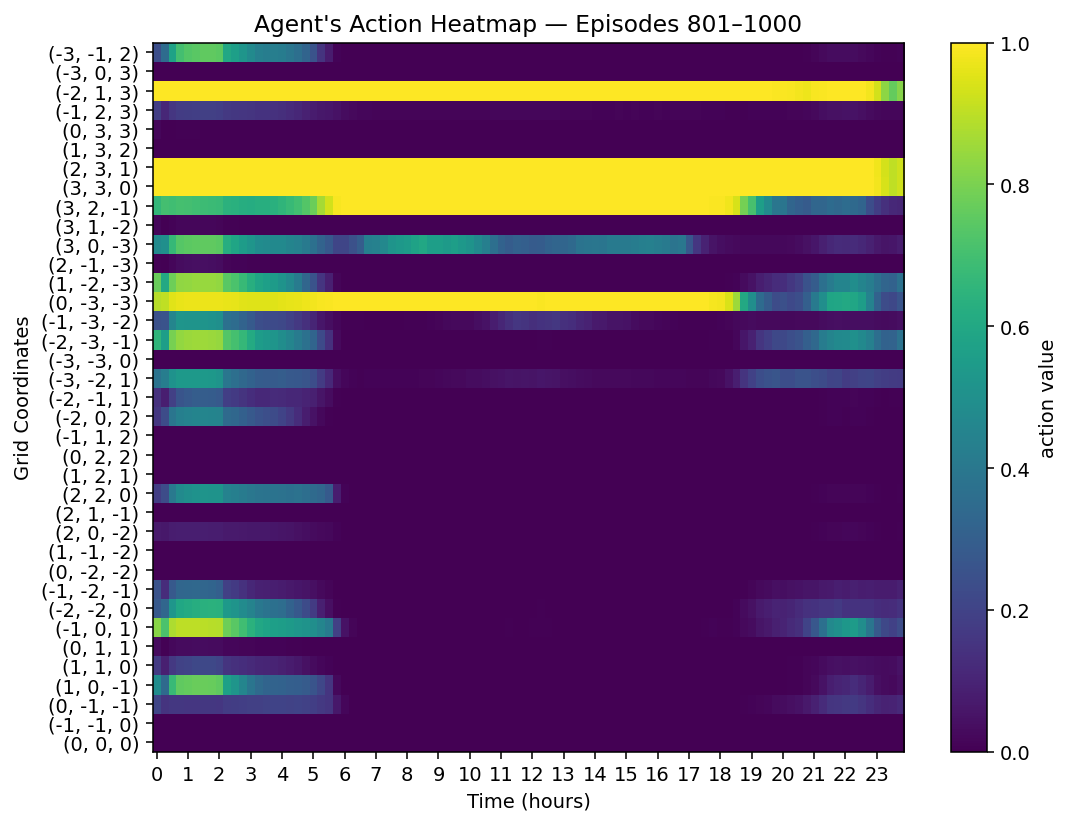}
        {\small \textbf{Episode} 801-1000}
    \end{minipage}

    \caption{Evolution of RL agent's policy}
    \label{fig:policy evolution}
\end{figure}

\subsubsection{Policy evolution and operational insights}

\begin{figure}[!htbp]
    \centering
    \includegraphics[width=1\linewidth]{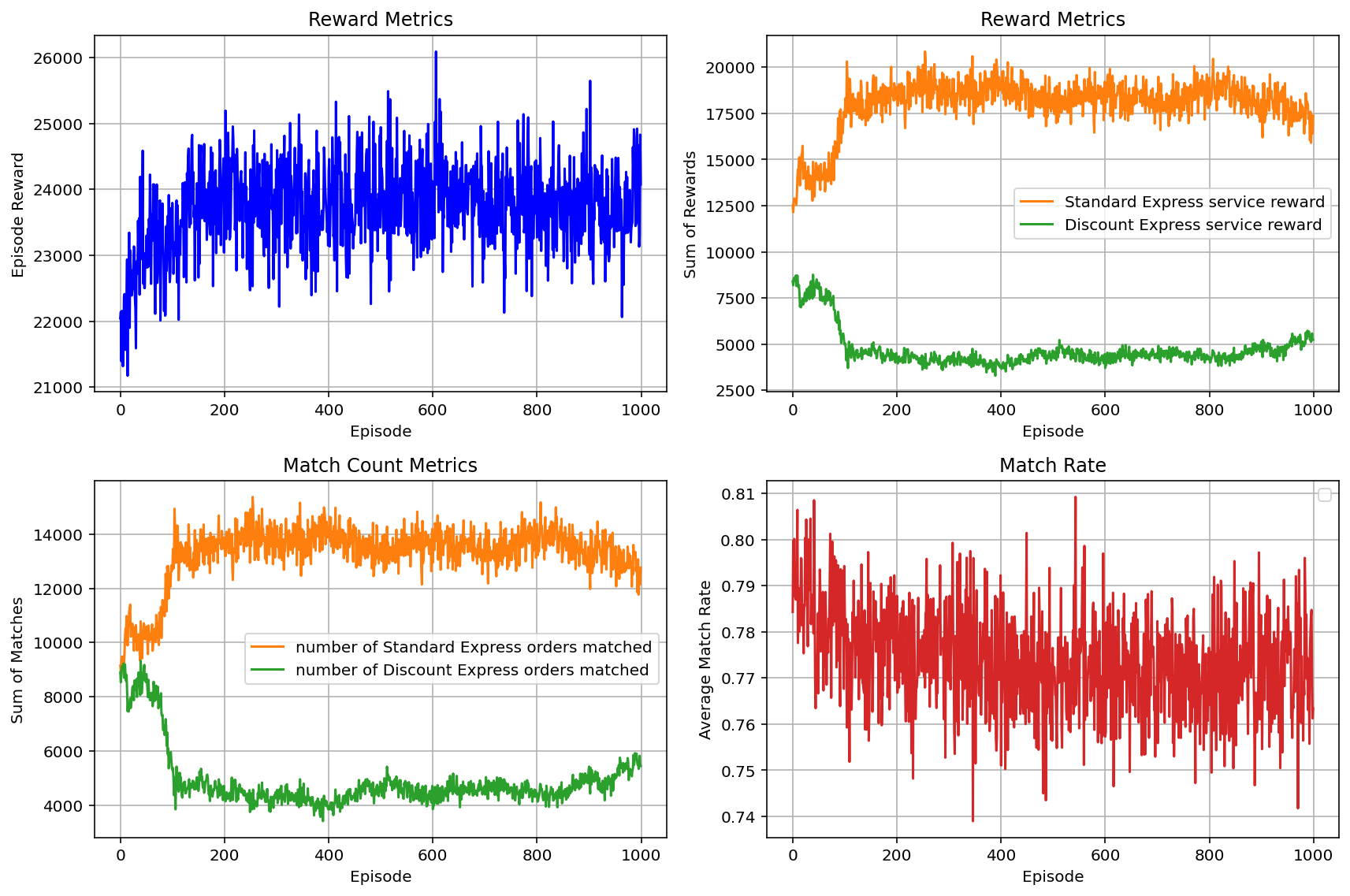}
    \caption{Curves of key operational metrics with the evolution of the actor policy}
    \label{fig: operational metrics curves}
\end{figure}

To further interpret how the proposed framework achieves these performance improvements, we conduct a detailed examination of the learned actor policy and its evolution during training. In Section~\ref{subsec: data processing}, Figure~\ref{fig: hourly trend of supply and demand} and Figure~\ref{fig: spatial distribution of demand} present the  temporal dynamics and spatial heterogeneity of passenger demands. Figure~\ref{fig:policy evolution} illustrate how the policy gradually learns to adapt to the spatial and temporal heterogeneity of passenger demand. In each heatmap of Figure~\ref{fig:policy evolution}, the horizontal axis represents time over a 24-hour period, while the vertical axis corresponds to the grid coordinates in the hexagonal network. The grids are ordered by their ring level, defined as $\text{ring level} = \max(|x|, |y|, |z|)$, which increases from bottom to top -- indicating progressively greater distance from the city center. The color intensity indicates the agent’s action value (i.e., the ratio of drivers' accepting Discount Express orders): brighter yellow denotes a higher ratio (closer to 1), whereas darker purple represents a lower ratio (closer to 0).
We also record several key operational metrics in the ride-hailing context (Figure~\ref{fig: operational metrics curves}), including the numbers of matched Standard Express and Discount Express orders, their corresponding service rewards, and the average match rate of idle drivers.

Specifically, in the early training stages (Episodes~1–400), the agent primarily captures the spatial demand pattern. As shown in the heatmaps, it quickly learns that the central grids -- where requests are dense -- can sustain sufficient matching rates without accepting Discount Express orders. In contrast, the peripheral grids with sparse demand consistently maintain high acceptance ratios for discount requests, to enhancing platform competitiveness in low-demand areas.

After approximately 400 training episodes, the policy begins to capture the temporal regularities of demand fluctuations. The agent learns that passenger requests are generally scarce during 0:00-6:00 and 22:00-24:00, and thus increases the acceptance of discount orders during these hours to attract more passengers and utilize idle drivers. During peak periods with high demand intensity (e.g., 6:00 to 22:00), the policy shifts toward rejecting discount requests in the central grids to preserve revenue from Standard Express orders.

In addition, the variations in key ride-hailing performance metrics further support the above observations (Figure~\ref{fig: operational metrics curves}). As training progresses, both the number of matched Discount Express orders and the corresponding service reward initially exhibit a sharp decline, reflecting the model’s early adjustment to disable drivers' discount order acceptance in high-demand central grids. Subsequently, these metrics rise gradually after around 400 episodes, coinciding with the policy’s increased acceptance of Discount Express requests during low-demand periods (0:00–6:00) and late-night hours after 22:00. Meanwhile, the overall driver match rate experiences a slight early decrease -- from approximately 79\% to 77\% -- as the system prioritizes higher-value Standard Express matches in central grids, followed by a gradual recovery.

Overall, this evolution demonstrates the adaptive capability of the proposed framework: it autonomously learns to differentiate between spatial and temporal demand contexts and adopts economically rational strategies—accepting Discount Express orders when demand is weak and rejecting them when demand is strong.

\subsubsection{Comparison of the early-episode training}

\begin{figure}[htbp]
    \centering
    \begin{minipage}{0.48\textwidth}
        \centering
        \includegraphics[width=\linewidth]{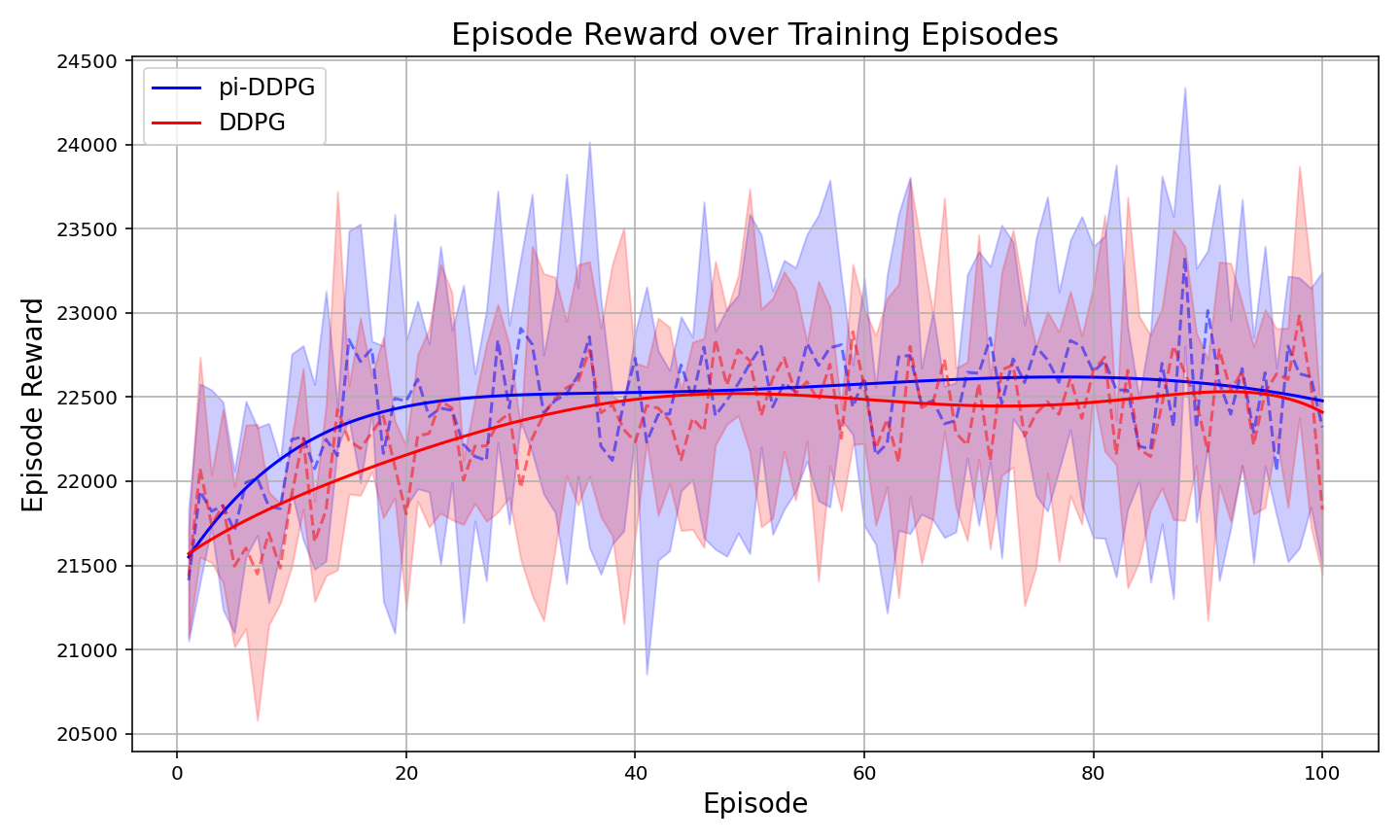}
    \end{minipage}
    \hfill
    \begin{minipage}{0.48\textwidth}
        \centering
        \includegraphics[width=\linewidth]{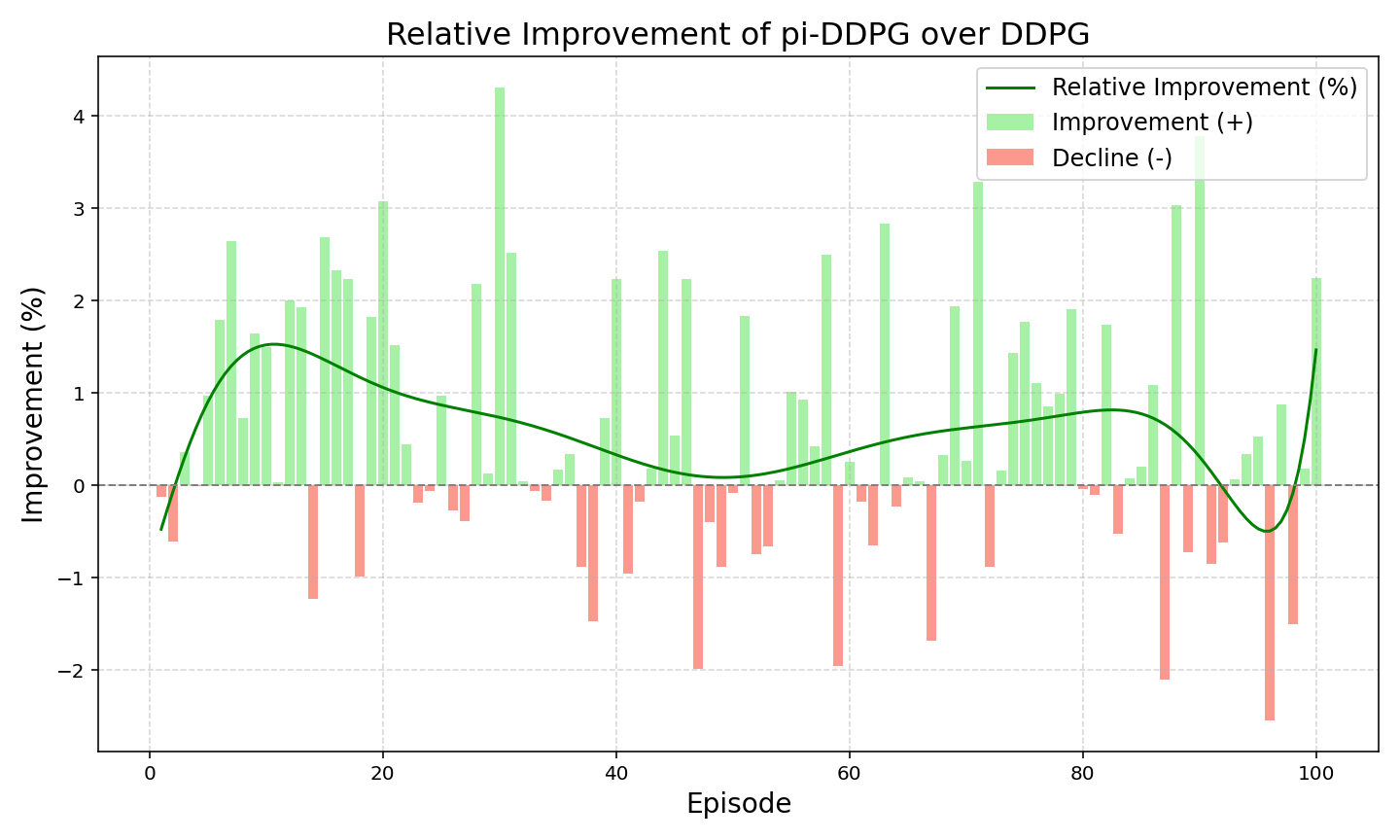}
    \end{minipage}

    \vspace{1em}

    % platform 2 title
    {\small (a) \textbf{Platform 1}}

    \begin{minipage}{0.48\textwidth}
        \centering
        \includegraphics[width=\linewidth]{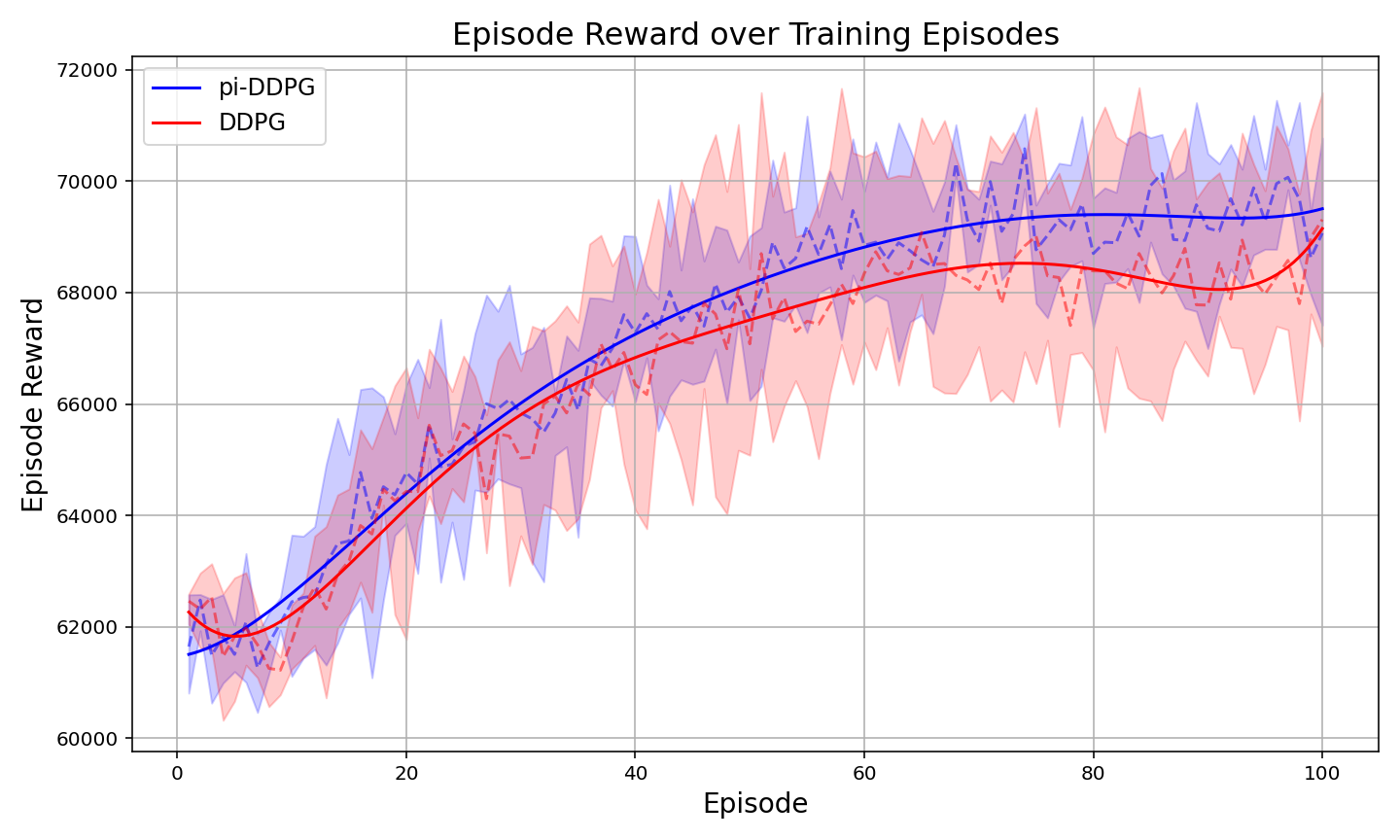}
    \end{minipage}
    \hfill
    \begin{minipage}{0.48\textwidth}
        \centering
        \includegraphics[width=\linewidth]{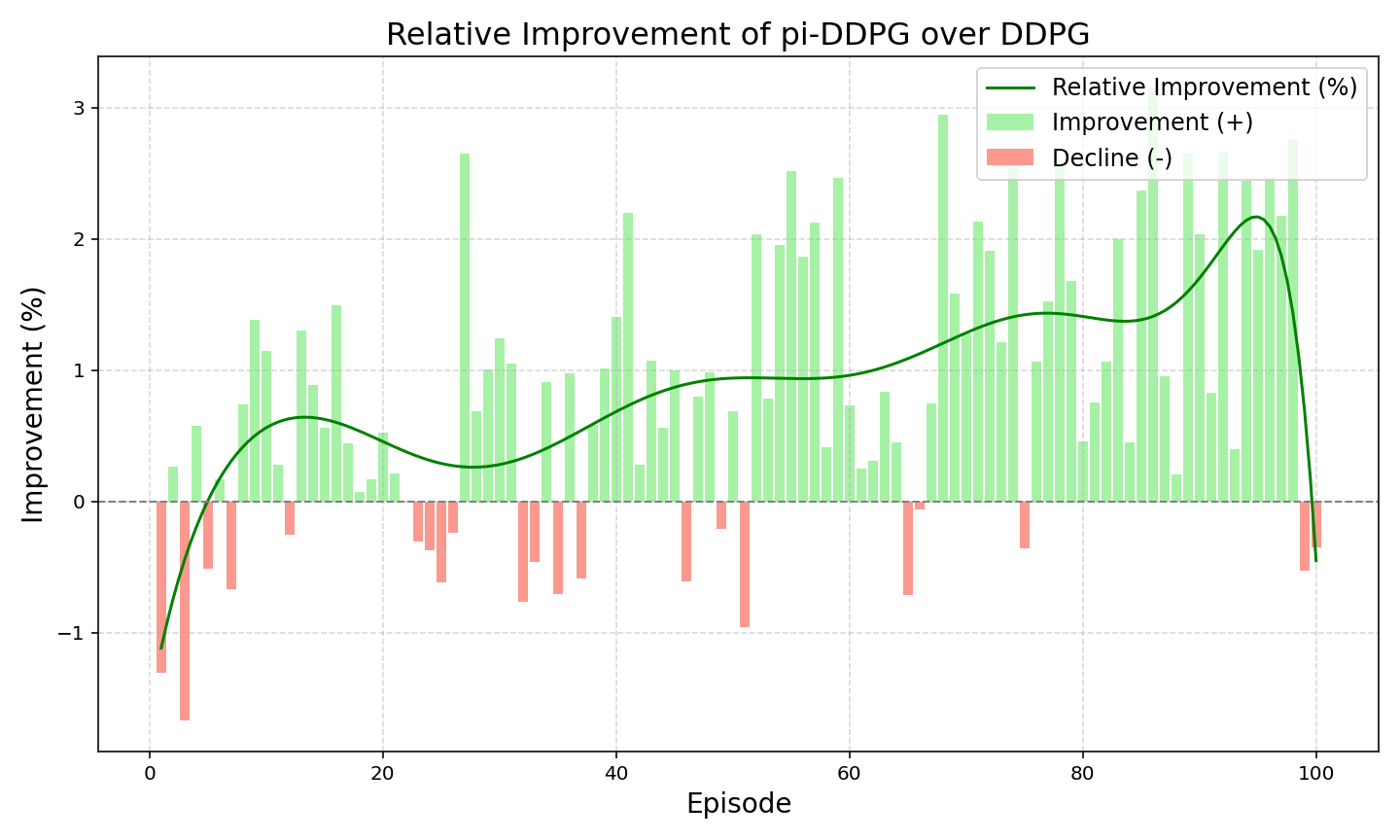}
    \end{minipage}

    \vspace{1em}

    % platform 5 title
    {\small (b) \textbf{Platform 2}}

    \begin{minipage}{0.48\textwidth}
        \centering
        \includegraphics[width=\linewidth]{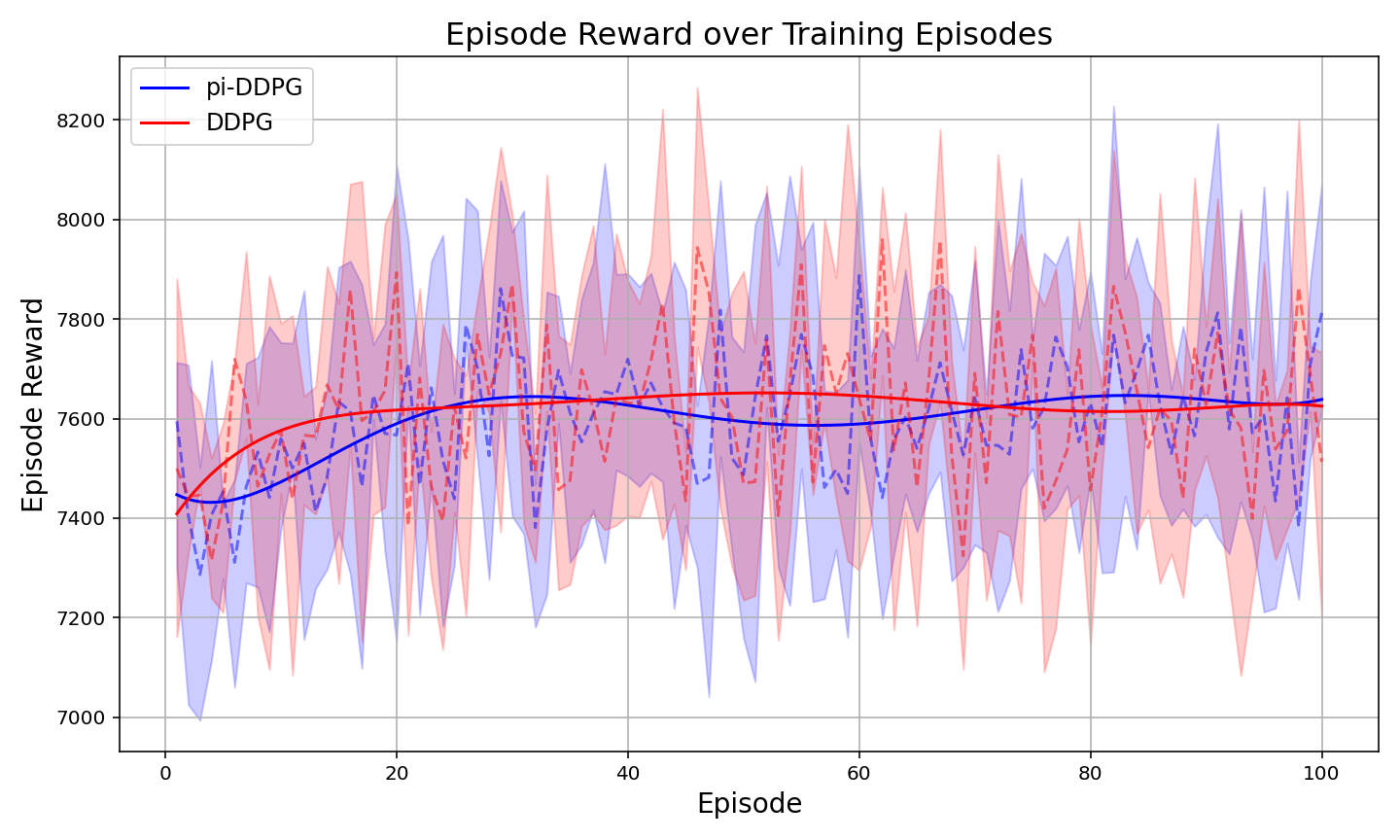}
    \end{minipage}
    \hfill
    \begin{minipage}{0.48\textwidth}
        \centering
        \includegraphics[width=\linewidth]{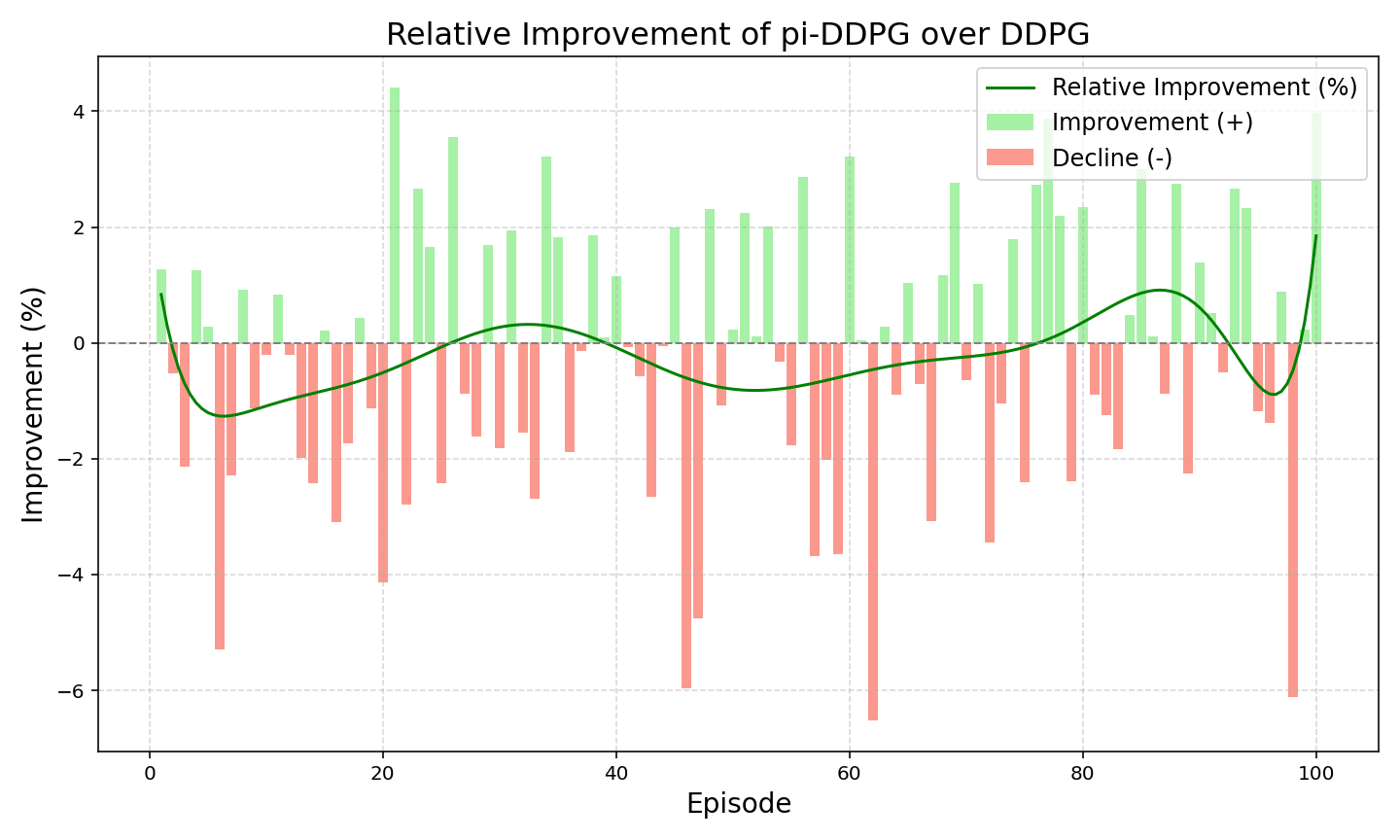}
    \end{minipage}

    \vspace{1em}

    % platform 5 title
    {\small (c) \textbf{Platform 5}}

    \caption{Comparison of early-episode training performance of DDPG and pi-DDPG across different platforms}
    \label{fig:comparison-early-training}
\end{figure}

Figure~\ref{fig:comparison-early-training} compares the early-stage learning performance (first 100 episodes) of the proposed pi-DDPG framework against the baseline DDPG across three representative platforms. For each individual platform, the left panel shows the episode reward trends over training, while the right panel illustrates the relative improvement percentage of pi-DDPG over DDPG at each episode. In the left panels, the solid curves represent the smoothed average episode rewards across 20 independent runs for each DDPG algorithm. And the shaded areas indicate the range from the 25th percentile to the 75th percentile of episode rewards observed among these 20 runs, reflecting the stability of each method during early training.

For \textbf{Platform~1}, pi-DDPG achieves a higher reward trajectory and faster convergence from the early episodes onward; after approximately 40 episodes, both algorithms reach comparable performance levels. For \textbf{Platform~2}, both algorithms exhibit smooth upward trends, but pi-DDPG shows faster convergence and maintains a consistent improvement of 0–5\% throughout from Episode 40 to 100. The lower variance of pi-DDPG indicates better learning stability.
\textbf{Platform~5}, which represents the most challenging operational context with lower baseline rewards, minimal difference between the two algorithms. In some episodes, pi-DDPG even performs up to 5\% worse than the baseline DDPG. This can be attributed to the unique characteristics of platform 5 in our simulation setting -- its total driver supply is the smallest among all platforms (Figure~\ref{fig: hourly trend of supply and demand}). As a result, the overall episode reward is lower due to limited service capacity. Moreover, the sparse spatial distribution of drivers often results in grids containing only a single driver, under which the actor’s continuous decision on the proportion of drivers accepting Discount Express orders effectively degenerates into a binary 0–1 outcome, making it difficult to distinguish between the two policies.

Overall, the pi-DDPG framework exhibits better early-stage online learning capability, for individual platforms with medium or large driver supply scales, which are critical in real-world applications with limited historical operational data of an emerging business mode and high exploration cost for a ride-hailing company.

%% file: data/section7_conclusion.tex
\section{Conclusion} \label{sec 7}

This study addresses a critical operational challenge for the individual platform in the competitive integrated platform, that is, how to dynamically determine the proportion of drivers accepting Discount Express orders to maximize the individual platform's profit and sustain competitive advantage. To improve the practicality in the online-fashion, this study proposes a novel pi-DDPG which extends the standard DDPG architecture by incorporating two key enhancements: a ConvLSTM-based spatiotemporal encoder and a refiner module that improves policy quality through localized Q-value guided optimization. The ConvLSTM component enables the agent to capture temporal and spatial dynamics of driver supply and customer demand, while the refiner serves as a lightweight correction mechanism that is especially effective during the early training phase. Extensive experiments conducted in a simulator calibrated with real-world data confirm that pi-DDPG outperforms the baseline DDPG in terms of early-stage episode rewards, convergence speed, and performance stability. However, these advantages become less pronounced in low-supply or highly sparse environments, where the policy learning struggles due to sharp 0–1 discretization in actions. While the refiner module enhances actor performance, it is not a replacement; the actor remains critical for generating valid initial actions and respo long-term policy learning. Together, the proposed enhancements offer a practical and data-driven framework for intelligent driver control under platform competition, with potential applications to other domains involving continuous action control.

Future work can proceed along two directions. First, the consideration of drivers' equity can be incorporated into the decision-making process by formulating a multi-objective reinforcement learning framework. The individual platform needs to balance between maximizing total service rewards and ensuring equity of order matching opportunities among drivers with accepting or rejecting discount order acceptance. Second, while this study focuses on a single platform, in practice, other individual platforms on the integrated platform may also adopt reinforcement learning strategies to optimize their policies. This would lead to a complex group dynamics, where individual platforms' policies co-evolve and impact each other. Addressing this challenge calls for further investigation into multi-agent reinforcement learning under a competitive environment, potentially enhancing the effectiveness and adaptability of the DDPG algorithm in real-world ride-hailing markets.

%% file: data/acknowledgements.tex
\section*{Acknowledgements}
%The research is supported in part by grants from National Natural Science Foundation of China (72371141).

The acknowledgements section will be completed after the peer-review process.

%% file: data/declaration.tex
\section*{Declaration of generative AI and AI-assisted technologies in the writing process}
        
During the preparation of this work the authors used ChatGPT 4o in order to improve language and help write \LaTeX. After using this tool/service, the authors reviewed and edited the content as needed and take full responsibility for the content of the publication.